\documentclass{article}

\usepackage[preprint]{neurips_2026}

\usepackage[utf8]{inputenc}
\usepackage[T1]{fontenc}
\usepackage{hyperref}
\usepackage{url}
\usepackage{booktabs}
\usepackage{amsfonts}
\usepackage{nicefrac}
\usepackage[expansion=false,protrusion=false]{microtype}
\usepackage{xcolor}

\usepackage{amsmath,amssymb,amsthm}
\usepackage{mathtools}
\usepackage{graphicx}
\usepackage{subcaption}
\usepackage{multirow}
\usepackage{algorithm}
\usepackage{algpseudocode}
\usepackage{siunitx}

\newcommand{\wmd}{\mathbf{w}_{\mathrm{LDA}}}
\newcommand{\wopt}{\mathbf{w}_{\mathrm{opt}}}
\newcommand{\wpca}{\mathbf{w}_{\mathrm{PC1}}}

\newcommand{\wmdp}[1]{\mathbf{w}_{\mathrm{LDA}}^{\mathrm{#1}}}
\newcommand{\woptp}[1]{\mathbf{w}_{\mathrm{opt}}^{\mathrm{#1}}}
\newcommand{\Lp}[1]{L_{\mathrm{#1}}}

\newcommand{\harm}{\mathcal{H}}
\newcommand{\norm}{\mathcal{N}}

\newcommand{\R}{\mathbb{R}}
\DeclareMathOperator*{\argmax}{arg\,max}

\title{Harmful Intent as a Geometrically Recoverable Feature of LLM Residual Streams}

\author{%
  Isaac Llorente-Saguer \\
  Independent Researcher\\
  \texttt{illorentes@proton.me} \\
}

\begin{document}
\maketitle

\begin{abstract}
Aligned language models refuse harmful instructions, but the
representations through which they recognise such instructions
are less well characterised than the behaviours they produce.
Harmful intent is linearly separable from residual-stream
activations across 12 models spanning four architectural
families (Qwen2.5, Qwen3.5, Llama-3.2, Gemma-3) and three
alignment variants (base, instruction-tuned, abliterated), with
parameter scales from 0.5B to 1.3B and a within-family
scale extension to 9B on Qwen3.5.
A direction fitted from 100 labelled examples per class via
Soft-AUC optimisation reaches mean effective AUROC 0.982 and
TPR@1\%FPR 0.797, generalises to three held-out harm
benchmarks and a hard-benign control, and matches its
instruction-tuned counterpart within $\pm 0.003$ AUROC in
abliterated variants from which the refusal mechanism has been removed.
The supervised strategies all exceed AUROC 0.96, but their
TPR@1\%FPR varies by more than ten times the AUROC gap;
a deployed 9B safety classifier shows the same pattern at
AUROC 0.94 and TPR 0.30, motivating low-FPR reporting as a
default in safety-adjacent detection evaluation.
Geometric measurements refine the picture.
The recovered direction is concentrated within each extraction
protocol but protocol-dependent across them: two pooling 
choices applied to the same chat-templated activations at the 
same residual-stream layer (max-pool over content tokens versus 
last-token at the post-instruction position) recover 
harm directions $73^\circ$ apart, and projecting one 
out leaves detection under either max-pool extraction 
essentially intact.
Probing identifies a protocol-specific direction rather than a
unique computational feature.
\end{abstract}

\section{Introduction}
\label{sec:introduction}

Aligned language models refuse harmful instructions reliably
enough that attention has shifted from whether they refuse to
how they fail to.
The mechanism by which these models recognise that an
instruction is harmful, as distinct from the mechanism by
which they act on that recognition, has received less empirical
attention than the behavioural surface.
This separation matters: a model that recognises harm but fails
to refuse it implies a different intervention from a model that
fails to recognise it at all, and the two failure modes cannot
be distinguished from output behaviour alone.

Recent work characterises the residual-stream geometry of
safety-relevant features.
\citet{arditi2024refusal} identified a direction in
instruction-tuned models that mediates refusal: ablating it
releases refusal, inserting it induces refusal on benign
prompts.
The procedure produces \emph{abliterated} model variants in
which refusal has been removed, and these variants
are publicly available across major model families.
\citet{zhao2025llms} showed through causal steering that
harmfulness and refusal are decodable as separable signals at
distinct token positions, mechanistically dissociable in three
instruction-tuned 7--8B models.
Together with broader representation-engineering
work~\citep{zou2023representation}, these results establish that
harmful intent is linearly decodable from transformer
activations and that recognition and refusal are separately
encoded features.

Three questions remain open.
First, does the harm-recognition signal generalise across
architectures and alignment interventions, or do different
families and alignment procedures produce qualitatively
different geometries?
Mechanistic evidence to date concentrates at a single
architectural family and alignment stage.
Second, what is the geometric relationship between
directions recovered under different extraction choices, and
between these directions and other safety-relevant axes such as
the dominant variance of benign-prompt activations?
Whether activation-probing recovers a unique computational
feature or a protocol-specific decoding axis has not been
systematically measured.
Third, how reliable is activation-based detection under realistic
deployment constraints, where false alarms must remain extremely
low and AUROC numbers can mask substantial operating-point
variation?

We evaluate six direction-fitting strategies on 12 models
spanning four architectural families (Qwen2.5, Qwen3.5,
Llama-3.2, Gemma-3) and three alignment variants (base,
instruction-tuned, abliterated), under a strict three-way data
split that separates fitting, validation, and evaluation at the
dataset level.
Two supervised projection strategies and one supervised
angular-deviation strategy each recover stable harm directions
from 100 labelled examples per class.
We characterise direction geometry within and across four
extraction protocols (combinations of max-pool vs.\ last-token
pooling and raw vs.\ chat-templated input), evaluate transfer
across datasets and alignment variants, extend to Qwen3.5 at
2B, 4B, and 9B parameters, and benchmark against four publicly
available safety classifiers.
Throughout we report low-FPR operating-point metrics alongside
AUROC.

\paragraph{Contributions.}
\begin{enumerate}
  \item \textbf{Stable linear recovery across architectures and
        alignment.} A mean-difference direction fitted from 100 
        examples per class achieves mean effective AUROC $0.982$ 
        across 12 models (Qwen2.5, Qwen3.5, Llama-3.2, Gemma-3; 
        base, instruct, abliterated; 0.5B--1.3B), with 
        within-family scale stability tested on Qwen3.5 up to 
        9B. Cross-variant transfer is AUROC-robust ($\geq 0.90$ 
        in every pair tested).

  \item \textbf{Detection persists through alignment 
        interventions.} AUROC differences between abliterated 
        variants and their instruction-tuned counterparts are 
        within $\pm 0.003$, indicating that the recognition 
        signal at content-token positions survives the  
        removal of the refusal direction from model 
        weights~\citep{arditi2024refusal}. This complements at 
        scale the mechanistic harm/refusal dissociation 
        reported by~\citet{zhao2025llms}.

  \item \textbf{Geometric characterisation of the harm 
        direction.} The recovered direction is concentrated 
        within each extraction protocol but protocol-dependent 
        across them: the directions recovered at content-token 
        positions and at the post-instruction position 
        ($\wmdp{lt/chat}$, the construction~\citet{arditi2024refusal} 
        use to identify the refusal direction) lie 
        $73^\circ \pm 7^\circ$ apart in the same residual-stream 
        layer; projecting $\wmdp{lt/chat}$ out leaves the 
        mean-difference vector under max-pool extraction 
        with $\geq 89\%$ of its norm intact. The recovered 
        direction is also nearly orthogonal to the leading 
        principal component of benign-prompt activations 
        ($82^\circ \pm 4^\circ$).

  \item \textbf{Operating-point evaluation.} TPR@1\%FPR varies 
        by more than ten times the AUROC gap among supervised 
        strategies that all exceed AUROC $0.96$. A deployed 
        $9$B safety classifier shows the same pattern at 
        AUROC $0.94$ / TPR $0.30$. We recommend low-FPR 
        metrics as a default alongside AUROC in safety-adjacent 
        detection.
\end{enumerate}

\section{Methods}
\label{sec:methods}

\subsection{Models, datasets, and notation}
\label{sec:setup}

We evaluate 12 models spanning four architectural families
(Qwen2.5~\citep{qwen2.5,qwen2}, Qwen3.5~\citep{qwen3.5},
Llama-3.2~\citep{dubey2024llama3herdmodels},
Gemma-3~\citep{gemma3_2025}) and three alignment variants per
family (base, instruction-tuned, abliterated;
Table~\ref{tab:model_ids} in Appendix~\ref{app:datasets}).
Abliterated variants are community-produced via the
weight-orthogonalisation procedure
of~\citet{arditi2024refusal} and sourced from HuggingFace under
the \texttt{huihui-ai} organisation.
The main analysis covers 0.5--1.3B parameters; we additionally
evaluate Qwen3.5 at 2B, 4B, and 9B.

\paragraph{Datasets and splits.}
Harmful prompts are drawn from
AdvBench~\citep{zou2023universal} (520 direct instructions),
HarmBench~\citep{mazeika2024harmbench} (200 standard behaviours),
and JailbreakBench~\citep{chao2024jailbreakbench} (100 harmful
goals); benign prompts from
Alpaca-Cleaned~\citep{taori2023alpaca} (general-purpose
instructions) and XSTest~\citep{rottger2024xstest} (250
hard-benign prompts that superficially resemble harmful queries).
Data is partitioned into three sample-disjoint sets at the
dataset level: a \emph{fit set} (100 harmful + 100 benign for
direction fitting, drawn from AdvBench and Alpaca only); a
\emph{validation set} (50 harmful + 50 benign for layer
selection, also from AdvBench and Alpaca); and an
\emph{evaluation set} that includes held-out AdvBench, Alpaca,
and the three sources withheld entirely from fitting (HarmBench,
JailbreakBench, XSTest).
This design provides both in-distribution evaluation (held-out
AdvBench vs.\ Alpaca) and dataset-level out-of-distribution
evaluation across three additional harm sources and one
additional benign source.
Per-source counts are in Appendix~\ref{app:datasets}.
Evaluation is single-turn, English, and on cleanly formatted prompts (Section~\ref{sec:discussion} discusses scope).

\paragraph{Prompt normalisation.}
All prompts are normalised by stripping terminal punctuation
(\texttt{[.,;:!?]}).
AdvBench harmful prompts uniformly terminate with a full stop
while Alpaca instructions typically do not; without this step, a trivial punctuation classifier nearly perfectly separates the two sources.

\paragraph{Activation extraction and protocols.}
Let $h_{\ell,t,d}$ denote dimension $d$ of the residual stream
at layer $\ell$, token position $t$, for a given prompt
$\mathbf{p}$.
We consider four extraction protocols, each producing a vector
in $\R^D$ at a chosen layer.
A protocol is identified by a (pooling, input-formatting) pair:
\emph{pooling} is either max-pool over content-token positions
(mp) or last-token (lt), and \emph{input-formatting} is either
the raw prompt (raw) or the prompt wrapped in the model's chat
template (chat).
We write the four combinations as
$\mathrm{mp/raw}$, $\mathrm{mp/chat}$, $\mathrm{lt/chat}$, and
$\mathrm{lt/raw}$.
For pooling protocol $P \in \{\mathrm{mp}, \mathrm{lt}\}$ and
input formatting $F \in \{\mathrm{raw}, \mathrm{chat}\}$, the
extraction operation is
$f^{P/F}_\ell(\mathbf{p})_d = \max_{t \in \mathcal{T}_F} 
h_{\ell,t,d}$ for $P = \mathrm{mp}$ (over the set 
$\mathcal{T}_F$ of token positions in the prompt under formatting 
$F$) and $f^{P/F}_\ell(\mathbf{p})_d = h_{\ell,t_\mathrm{last},d}$
for $P = \mathrm{lt}$ (the residual stream at the final token of
the formatted prompt).

The main analysis
(Sections~\ref{sec:main_results}--\ref{sec:dissociation_results},
Section~\ref{sec:generalization}, Section~\ref{sec:classifier_comparison_results})
uses $\mathrm{mp/raw}$ as the single protocol applied across all
12 models, since base models lack chat templates and a
privileged decision position, which precludes a $\mathrm{chat}$-
or $\mathrm{lt}$-based protocol as a cross-variant choice.
The geometric experiments in Section~\ref{sec:geometry} compare
directions across multiple protocols on the four
instruction-tuned models, where chat templates are well-defined
and the post-instruction position is meaningful.
The chat-template robustness comparison
(Section~\ref{sec:chat_template}) uses $\mathrm{lt/raw}$ and
$\mathrm{lt/chat}$.

Activations are extracted via forward hooks on each transformer
block's output and cached to disk per model.
Direction fitting and scoring are pure NumPy operations on the
cached tensors.
All experiments are seeded globally (Python, NumPy, PyTorch) at
seed~42.

\subsection{Direction strategies}
\label{sec:directions}

We evaluate six direction strategies, falling into two
families: \emph{projection-based} methods that score prompts by
signed projection onto a direction, and \emph{angle-based}
methods (marked $\theta$) that use unsigned angular deviation.
Each direction is fitted under one extraction protocol and
inherits a protocol superscript marking the (pooling,
input-formatting) pair.
For a chosen layer $\ell$ and protocol $P/F$, the fit set
$\mathcal{F}_\harm = \{\mathbf{x}_i^+\}_{i=1}^{n_+}$ of harmful
activations and $\mathcal{F}_\norm = \{\mathbf{x}_j^-\}_{j=1}^{n_-}$
of benign activations are extracted by
$f^{P/F}_\ell(\cdot)$.
We seek a unit-norm direction $\mathbf{w}^{P/F} \in \R^D$ whose
scalar projection
$s(\mathbf{p}) = f^{P/F}_\ell(\mathbf{p}) \cdot \mathbf{w}^{P/F}$
provides a harmful-intent score under that protocol.

\paragraph{Mean difference, $\wmdp{P/F}$.}
\begin{equation}
  \wmdp{P/F} =
    \frac{\hat{\mu}^{P/F}_\harm - \hat{\mu}^{P/F}_\norm}
         {\|\hat{\mu}^{P/F}_\harm - \hat{\mu}^{P/F}_\norm\|},
  \label{eq:mean_diff}
\end{equation}
where $\hat{\mu}^{P/F}_\harm$ and $\hat{\mu}^{P/F}_\norm$ are
the sample class means of activations extracted under protocol
$P/F$.
Under equal spherical within-class covariances, this is the
Fisher Linear Discriminant direction.
Fitting cost is $O((n_+ + n_-)\,D)$.
The same arithmetic underlies the refusal direction
of~\citet{arditi2024refusal}, which in our notation is
$\wmdp{lt/chat}$:
the two are mean-difference directions distinguished by
extraction protocol (max-pool over content tokens of raw input
in the main analysis vs.\ last-token of chat-templated input in
Arditi's procedure) and by validation criterion (held-out
discrimination AUROC vs.\ causal effect on refusal generation).

\paragraph{Soft-AUC optimised, $\woptp{P/F}$.}
We maximise a soft-AUC surrogate~\citep{calders2007efficient}
via Riemannian gradient ascent on the unit sphere:
\begin{equation}
  \hat{U}(\mathbf{w}) =
    \frac{1}{n_+ n_-}
    \sum_{i \in +,\, j \in -}
    \sigma\!\left(\mathbf{w}^\top(\mathbf{x}_i - \mathbf{x}_j)\right),
  \label{eq:soft_auc}
\end{equation}
with $\sigma$ the logistic sigmoid, and warm start from the
protocol's $\wmdp{P/F}$.
Implementation details are in Appendix~\ref{app:softauc}.

\paragraph{$\theta$ two-class optimised.}
The same Soft-AUC procedure as $\woptp{P/F}$, applied to
angular deviation scores
$s(\mathbf{p}) = \arccos(\bar{\mathbf{x}}(\mathbf{p}) \cdot
\mathbf{w})$ rather than projection scores, where
$\bar{(\cdot)}$ denotes $\ell_2$ normalisation.
Warm-started from the $\theta$-normative centroid direction,
extending the angular deviation framework of
\citep{llorente2025theta, llorente2026geometry} to supervised
class discrimination.

\paragraph{Zero-shot baselines.}
\emph{PC1 (normative), $\wpca$:} the leading principal component
of $\mathcal{F}_\norm$.
\emph{$\theta$-normative:} the unsigned angular deviation from
the normalised normative centroid $\mu_\norm / \|\mu_\norm\|$,
following~\citep{llorente2026geometry}.
\emph{Random:} a uniformly random unit vector at seed~42.
\emph{Perplexity:} the prompt's mean per-token negative
log-likelihood under the model whose activations are
probed~\citep{zhu2023autodan}.

\subsection{Scoring, metrics, and layer selection}
\label{sec:scoring}

Detection scores are signed projections (projection-based
strategies) or unsigned angular deviations (angle-based
strategies), as defined in Section~\ref{sec:directions}.
Unsupervised strategies (PC1, $\theta$-normative, random) can
produce scores negatively correlated with harm; we apply a sign
correction $s \leftarrow -s$ when the empirical AUROC on the
evaluation set is below $0.5$ and report the
\emph{effective AUROC}
$\max(\mathrm{AUROC},\, 1 - \mathrm{AUROC})$.
The supervised strategies $\wmd$, $\wopt$, and $\theta$
two-class produce sign-correct scores by construction.

We report two primary metrics.
\emph{Effective AUROC} summarises discriminability across all
thresholds.
\emph{TPR@1\%FPR} is the true-positive rate at a 1\%
false-positive rate, estimated from the empirical ROC curve by
linear interpolation at $\mathrm{FPR}=0.01$.
For each model and strategy we compute stratified bootstrap
95\% confidence intervals (1{,}000 resamples, stratified by
evaluation source).

The operating layer is selected on the validation set
\emph{per protocol}: for each layer $\ell$, $\wmdp{P/F}$ is
fitted on the fit set extracted under protocol $P/F$ and scored
on the validation set, and we take
$\ell^*_{P/F} = \argmax_\ell \mathrm{AUROC}_\mathrm{val}(\ell)$.
This yields $\Lp{mp/raw}$ for the main analysis,
$\Lp{lt/chat}$ for Arditi's protocol comparison and the
projection-and-refit experiments, and $\Lp{mp/chat}$,
$\Lp{lt/raw}$ where the corresponding protocols are tested
(Section~\ref{sec:geometry},
Section~\ref{sec:chat_template}).
At each protocol's selected layer, all strategies trained under
that protocol are fitted and evaluated, ensuring direction
angles within a protocol are comparable across methods at the
same residual-stream layer.

\subsection{Comparison with dedicated safety classifiers}
\label{sec:baseline_classifiers}

We benchmark four publicly available safety classifiers on the
same evaluation set: Llama Guard~3
(8B)~\citep{dubey2024llama3herdmodels}, WildGuard
(7B)~\citep{han2024wildguard}, ShieldGemma
(9B)~\citep{zeng2024shieldgemma}, and Latent Guard, a probe
over model activations ~\citep{zhao2025llms}.
WildGuard is fine-tuned from Mistral-7B-v0.3 on
WildGuardTrain~\citep{han2024wildguard};
Llama Guard~3 is trained on hh-rlhf-derived prompts with
Llama 2 and Llama 3 generations, supplemented with human and
synthetic safety data~\citep{dubey2024llama3herdmodels}.
For each classifier we report effective AUROC, TPR@1\%FPR, and
end-to-end inference latency on a single NVIDIA RTX~3090.

\section{Results}
\label{sec:results}

\subsection{Detection across architectures and alignment variants}
\label{sec:main_results}

A linear direction fitted from 100 examples per class achieves 
near-ceiling discrimination across all 12 models. 
$\woptp{mp/raw}$ achieves mean effective AUROC 
$0.982 \pm 0.006$ (min 0.970, Gemma-3-base; max 0.991, 
Llama-3.2-1B-abliterated) and mean TPR@1\%FPR 
$0.797 \pm 0.056$. $\wmdp{mp/raw}$ reaches AUROC 
$0.975 \pm 0.010$ and TPR $0.706 \pm 0.103$ at sub-millisecond 
fitting cost. The angle-based supervised strategy 
$\theta$~two-class achieves AUROC $0.962 \pm 0.027$ and TPR 
$0.607 \pm 0.152$. Zero-shot strategies and the perplexity 
baseline reach AUROC below 0.78 and TPR below 0.05 on every 
model (Appendix~\ref{app:full_results}).

\begin{figure}[ht]
\centering
\includegraphics[width=0.95\linewidth]{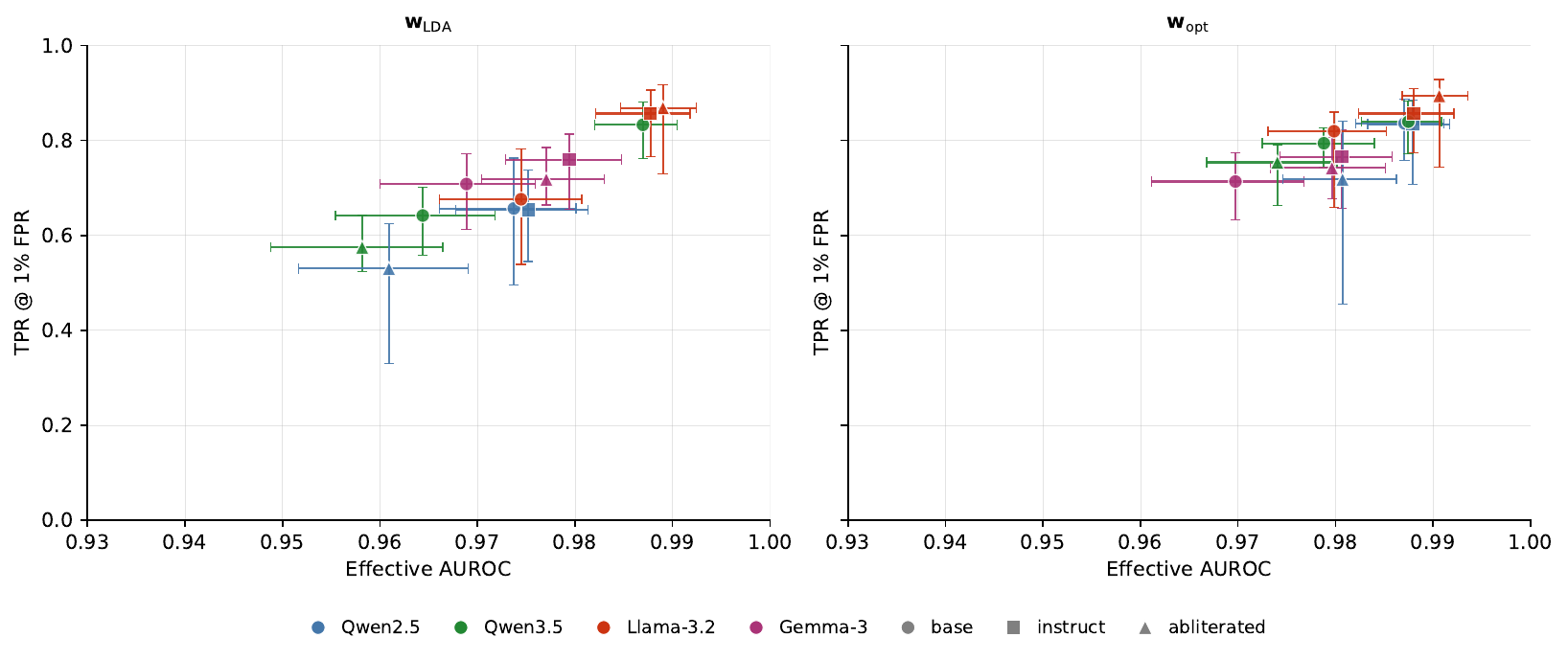}
\caption{Effective AUROC versus TPR@1\%FPR for 
$\wmdp{mp/raw}$ (left) and $\woptp{mp/raw}$ (right) 
across 12 models, with stratified bootstrap 95\% CIs (1{,}000 
resamples). Each point is one model; colour indicates 
architecture family, shape indicates alignment variant. Within 
the AUROC 0.96+ regime, TPR varies by up to 0.34 across models.}
\label{fig:auroc_tpr}
\end{figure}

\paragraph{AUROC and TPR@1\%FPR diverge in the near-ceiling regime.}
All supervised strategies exceed AUROC 0.96, so AUROC alone 
makes them appear interchangeable. TPR@1\%FPR disagrees: the 
gap between $\woptp{mp/raw}$ (0.797) and 
$\wmdp{mp/raw}$ (0.706) is over an order of magnitude 
larger than the 0.007 AUROC difference 
(Figure~\ref{fig:auroc_tpr}). The most extreme case is the 
zero-shot PC1 baseline at AUROC 0.665 with TPR below 0.05 on 
every model.

\subsection{Detection persists through alignment interventions}
\label{sec:dissociation_results}

The maximum within-family AUROC range across base, instruct, and
abliterated variants is 0.013 for $\woptp{mp/raw}$ and 
0.029 for $\wmdp{mp/raw}$ 
(Appendix~\ref{app:full_results}). Abliterated variants, from 
which the refusal direction has been removed via the 
weight-orthogonalisation procedure 
of~\citet{arditi2024refusal}, match their instruction-tuned 
counterparts within $\pm 0.003$ AUROC.

\paragraph{Cross-variant direction transfer.}
A direction fitted on the base model transfers to
instruction-tuned and abliterated variants with at most 0.010
AUROC degradation in three of four families
(Table~\ref{tab:cross_variant}). For Qwen2.5, Qwen3.5, and 
Llama-3.2, base-to-instruct angular shifts range from 
$17^\circ$ to $39^\circ$ and AUROC transfer remains within 
0.010 of own-direction performance. Gemma-3 is the exception: 
alignment rotates the harm direction by $73^\circ$, 
base-to-instruct AUROC drops by 0.057, and TPR collapses from 
0.751 to 0.175. Each Gemma-3 variant's own direction recovers 
AUROC above 0.969. Across all four families, 
instruct-to-abliterated transfer is substantially cleaner than 
base-to-instruct (angles $11^\circ$ to $42^\circ$, AUROC 
degradation $\leq 0.003$). Full $3 \times 3$ transfer matrices 
including the asymmetric Gemma-3 case are reported in 
Appendix~\ref{app:cross_variant}.

\begin{table}[ht]
\centering\small
\caption{Cross-variant direction transfer summary. For each 
  family, the base model's $\wmdp{mp/raw}$ direction is 
  applied to instruction-tuned and abliterated variants at the 
  base model's validation-selected layer. Angles: pairwise 
  unsigned angles between each pair's $\wmdp{mp/raw}$ 
  directions (B: base, I: instruct, A: abliterated). Each 
  metric cell shows the transferred value with the delta 
  relative to the target variant's own direction in parentheses 
  (negative = transfer underperforms own). Full $3 \times 3$ 
  matrices in Appendix~\ref{app:cross_variant}.}
\label{tab:cross_variant}
\begin{tabular}{lccc cc cc}
\toprule
& \multicolumn{3}{c}{Angles} & \multicolumn{2}{c}{Instruct} & \multicolumn{2}{c}{Abliterated} \\
\cmidrule(lr){2-4} \cmidrule(lr){5-6} \cmidrule(lr){7-8}
Family & B$\leftrightarrow$I & B$\leftrightarrow$A & I$\leftrightarrow$A & AUROC & TPR & AUROC & TPR \\
\midrule
Qwen2.5   & 18$^\circ$ & 29$^\circ$ & 21$^\circ$ & 0.973 ($-0.002$) & 0.597 ($-0.046$) & 0.958 ($-0.003$) & 0.479 ($-0.031$) \\
Qwen3.5   & 17$^\circ$ & 20$^\circ$ & 11$^\circ$ & 0.988 ($+0.001$) & 0.815 ($-0.006$) & 0.986 ($+0.002$) & 0.767 ($-0.018$) \\
Llama-3.2 & 39$^\circ$ & 52$^\circ$ & 42$^\circ$ & 0.978 ($-0.010$) & 0.754 ($-0.073$) & 0.981 ($-0.007$) & 0.733 ($+0.031$) \\
Gemma-3   & 73$^\circ$ & 74$^\circ$ & 26$^\circ$ & 0.922 ($-0.057$) & 0.175 ($-0.576$) & 0.900 ($-0.077$) & 0.100 ($-0.612$) \\
\bottomrule
\end{tabular}
\end{table}

\subsection{Geometric structure of the harm direction}
\label{sec:geometry}

\paragraph{Within-protocol angles.}
The harm direction is nearly orthogonal to the dominant variance
of benign-prompt activations 
(Appendix~\ref{app:angles}). At $\Lp{mp/raw}$ across the 12 
models, the mean angle between $\wmdp{mp/raw}$ and the 
leading principal component of benign activations is 
$81.7^\circ \pm 4.1^\circ$. The 
$\wmdp{mp/raw}$--$\woptp{mp/raw}$ angle is 
$9.1^\circ \pm 8.8^\circ$: the two strategies recover 
near-identical directions. The angle between the normative and 
harmful first principal components varies from $39^\circ$ 
(Llama-3.2) to $78^\circ$ (Qwen3.5); a single mean-difference 
algorithm nonetheless recovers near-optimal directions across 
this heterogeneous set.

\paragraph{The angular strategy is geometrically distinct.}
At $\Lp{mp/raw}$, the $\theta$~two-class direction lies at 
$73.4^\circ \pm 12.3^\circ$ from $\wmdp{mp/raw}$. Across 
the three Qwen2.5 variants, $\wmdp{mp/raw}$ and 
$\woptp{mp/raw}$ degrade to AUROC below 0.80 across layers 
2--20 (a "valley" profile), while $\theta$~two-class sustains 
AUROC above 0.95 throughout 
(Appendix~\ref{app:layer_profiles}). 
TPR@1\%FPR is lower for the angular strategy than for 
$\woptp{mp/raw}$ (0.607 vs.\ 0.797), so it complements 
rather than replaces projection-based detection.

\paragraph{Relationship to the refusal direction.}
On the four instruction-tuned models we compare 
$\woptp{mp/chat}$ against $\wmdp{lt/chat}$, the refusal 
direction of~\citet{arditi2024refusal}. Both are mean-difference 
constructions on chat-templated activations, distinguished by 
pooling. The unsigned angle is 
$72.6^\circ \pm 6.7^\circ$ at $\Lp{mp/chat}$ and 
$74.9^\circ \pm 9.3^\circ$ at $\Lp{lt/chat}$, near-orthogonal 
under either layer choice (Table~\ref{tab:arditi_geometry}). 
Cross-protocol AUROC remains in $[0.91, 0.99]$: both directions 
retain near-ceiling discriminability under the other protocol's 
extraction, despite the geometric non-collinearity.

\begin{table}[ht]
\centering\small
\caption{Geometric relationship between $\woptp{mp/chat}$ and 
$\wmdp{lt/chat}$ on the four instruction-tuned models. 
\textbf{Protocols.} \emph{$\mathrm{mp/chat}$}: max-pool over 
content-token positions of chat-templated input. 
\emph{$\mathrm{lt/chat}$}: last-token of chat-templated input, 
the protocol used by~\citet{arditi2024refusal} to identify the 
refusal direction. \textbf{Layer column.} 
$\Lp{mp/chat}\,/\,\Lp{lt/chat}$, the validation-selected layer 
under each protocol; a single value indicates the two protocols 
selected the same layer. \textbf{Angle column.} Unsigned angle 
between $\woptp{mp/chat}$ and $\wmdp{lt/chat}$ in 
residual-stream space, shown as 
angle@$\Lp{mp/chat}\,/\,$angle@$\Lp{lt/chat}$ when the layers 
differ, and as a single value otherwise. 
\textbf{AUROC columns.} \emph{own}: direction scored on its own 
protocol's evaluation activations at its own 
validation-selected layer. \emph{cross}: direction scored on 
the other protocol's evaluation activations at the other 
direction's validation-selected layer. AUROC with stratified 
bootstrap 95\% CIs (1{,}000 resamples; CIs in 
Appendix~\ref{app:projection_test}).}
\label{tab:arditi_geometry}
\begin{tabular}{lcc cc cc}
\toprule
& & & \multicolumn{2}{c}{$\woptp{mp/chat}$ AUROC} & \multicolumn{2}{c}{$\wmdp{lt/chat}$ AUROC} \\
\cmidrule(lr){4-5} \cmidrule(lr){6-7}
Model (instruct) & Layer & Angle & own & cross & own & cross \\
\midrule
Qwen2.5-0.5B & 20\,/\,12 & $77.9^\circ$\,/\,$84.7^\circ$ & 0.982 & 0.966 & 0.990 & 0.906 \\
Qwen3.5-0.8B & 18 & $72.2^\circ$ & 0.990 & 0.975 & 0.986 & 0.948 \\
Llama-3.2-1B &  7\,/\,11 & $77.0^\circ$\,/\,$79.3^\circ$ & 0.992 & 0.996 & 0.998 & 0.921 \\
Gemma-3-1B & 15 & $63.3^\circ$ & 0.996 & 0.994 & 0.989 & 0.985 \\
\midrule
\textit{Mean} & & $72.6^\circ$\,/\,$74.9^\circ$ & 0.990 & 0.983 & 0.991 & 0.940 \\
\bottomrule
\end{tabular}
\end{table}

\paragraph{Within-protocol concentration of the harm signal.}
We test how concentrated the harm signal is along the recovered 
direction by projecting it out and refitting from scratch 
(Appendix~\ref{app:self_projection}). Under $\mathrm{mp/raw}$ 
extraction at $\Lp{mp/raw}$, refitting $\woptp{mp/raw}$ 
in the subspace orthogonal to $\wmdp{mp/raw}$ drops AUROC 
substantially: mean $0.598$ across the four instruction-tuned 
models, with Qwen2.5 and Qwen3.5 falling to near-chance 
($0.516$ and $0.540$ respectively). Under $\mathrm{lt/chat}$ 
extraction at $\Lp{lt/chat}$, the analogous procedure 
(refitting in the subspace orthogonal to $\wmdp{lt/chat}$) 
retains more signal: mean $0.780$, range $0.719$--$0.813$. The 
harm signal at the operating layer of each protocol is 
concentrated rather than spread across many non-collinear 
directions.

\paragraph{Cross-protocol non-collinearity.}
At $\Lp{lt/chat}$, we project $\wmdp{lt/chat}$ out of activations 
and refit under each of three protocols 
(Appendix~\ref{app:cross_protocol_projection}). Under 
$\mathrm{mp/raw}$ extraction, AUROC change is at most $0.010$ across the four instruction-tuned models 
(mean $-0.002$), and the mean-difference vector under 
$\mathrm{mp/raw}$ retains $\geq 99\%$ of its norm. Under 
$\mathrm{mp/chat}$ extraction, the change is also small in 
absolute value (mean $+0.006$, max $+0.020$ on Qwen2.5), and 
the mean-difference vector retains $89$--$98\%$ of its norm. 
Under $\mathrm{lt/chat}$ extraction (Arditi's own protocol), 
the projection drops AUROC by $0.19$--$0.40$, with 
Llama-3.2-1B-Instruct showing the largest drop ($0.999$ to 
$0.601$). The direction $\wmdp{lt/chat}$ is near-orthogonal to 
the mean-difference directions recovered by either max-pool 
extraction at the same residual-stream layer.

\subsection{Generalisation across distributions and scale}
\label{sec:generalization}

\paragraph{Cross-dataset transfer.}
Directions fitted on AdvBench transfer to held-out HarmBench 
and JailbreakBench without retraining (Table~\ref{tab:ood};
per-model breakdown in Appendix~\ref{app:ood}). The worst 
single-model out-of-distribution cell is 0.941 (Gemma-3-base, 
JailbreakBench vs.\ Alpaca); the lowest model-averaged AUROC 
against XSTest is 0.975.

\begin{table}[ht]
\centering\small
\caption{Out-of-distribution AUROC for $\woptp{mp/raw}$ 
  averaged across 12 models. Rows: harmful source. Columns: 
  benign source. AdvBench is used for fitting; all other 
  sources are held out. Per-model values in 
  Appendix~\ref{app:ood}.}
\label{tab:ood}
\begin{tabular}{lccc}
\toprule
Harmful source  & vs Alpaca & vs XSTest & min \\
\midrule
AdvBench         & 0.992 & 0.998 & 0.992 \\
HarmBench        & 0.961 & 0.979 & 0.961 \\
JailbreakBench   & 0.962 & 0.975 & 0.962 \\
\bottomrule
\end{tabular}
\end{table}

\paragraph{Scale stability.}
On Qwen3.5 at 2B, 4B, and 9B (Table~\ref{tab:transfer_scale}, 
Appendix~\ref{app:scale}), AUROC remains in 0.98--0.99, 
cross-variant transfer stays within 0.008 AUROC of own-direction 
performance. In TPR@1\%FPR,
the 4B model drops up to 0.094, but the rest of the models stay within 0.018.

\paragraph{Sample efficiency.}
$\wmdp{mp/raw}$ plateaus at $n_{\mathrm{fit}} = 50$ 
examples per class; $\woptp{mp/raw}$ continues to improve 
through $n = 100$, gaining $+0.093$ TPR@1\%FPR over 
$\wmdp{mp/raw}$ at the operating point 
(Appendix~\ref{app:sample_efficiency}).

\paragraph{Chat-template robustness.}
\label{sec:chat_template}
The main analysis uses $\mathrm{mp/raw}$ extraction so that a 
single protocol applies across base, instruction-tuned, and 
abliterated variants. We test an alternative 
protocol on the four instruction-tuned models (one per family): 
$\mathrm{lt/chat}$, last-token of chat-templated input. All 
four models recover stronger directions under this protocol 
(Table~\ref{tab:chat_template}): mean AUROC reaches 0.993, 
mean TPR@1\%FPR reaches 0.908. The improvement decomposes into 
two effects. Within last-token pooling, switching from raw to 
chat-templated input contributes $+0.215$ TPR (the 
$\mathrm{lt/raw}$ baseline averages 0.693). Compared to the 
main-analysis $\mathrm{mp/raw}$ protocol on the same four 
models (mean TPR 0.813, computed from 
Tables~\ref{tab:full_a}--\ref{tab:full_b}), the protocol switch 
to $\mathrm{lt/chat}$ contributes $+0.095$ TPR overall.

\begin{table}[ht]
\centering\small
\caption{Chat-template robustness on the four instruction-tuned 
models, last-token extraction. \textbf{Protocols.} 
\emph{$\mathrm{lt/raw}$}: last-token extraction from raw 
prompts at the per-model validation-selected layer 
$\Lp{lt/raw}$. \emph{$\mathrm{lt/chat}$}: last-token 
extraction from chat-templated input at the per-model 
validation-selected layer $\Lp{lt/chat}$. 
\textbf{Layer columns.} The selected layer under each protocol. 
\textbf{$\Delta$ columns.} Change relative to the 
$\mathrm{lt/raw}$ baseline 
($\Delta = \mathrm{lt/chat} - \mathrm{lt/raw}$). All metrics are 
$\woptp{P/F}$ AUROC and TPR@1\%FPR with stratified bootstrap 
95\% CIs (1{,}000 resamples; per-model CIs in 
Appendix~\ref{app:full_results}).}
\label{tab:chat_template}
\begin{tabular}{lcccccccc}
\toprule
 & \multicolumn{2}{c}{Layer} & \multicolumn{2}{c}{AUROC} & \multicolumn{2}{c}{TPR@1\%FPR} & $\Delta$ AUROC & $\Delta$ TPR \\
\cmidrule(lr){2-3} \cmidrule(lr){4-5} \cmidrule(lr){6-7}
Model & $\Lp{lt/raw}$ & $\Lp{lt/chat}$ & lt/raw & lt/chat & lt/raw & lt/chat & & \\
\midrule
Qwen2.5-0.5B-Instruct & 14 & 12 & 0.977 & 0.991 & 0.755 & 0.896 & +0.014 & +0.140 \\
Qwen3.5-0.8B & 16 & 18 & 0.958 & 0.988 & 0.557 & 0.858 & +0.031 & +0.301 \\
Llama-3.2-1B-Instruct & 6 & 11 & 0.979 & 0.999 & 0.745 & 0.960 & +0.020 & +0.215 \\
Gemma-3-1B-it & 16 & 15 & 0.974 & 0.994 & 0.717 & 0.919 & +0.019 & +0.202 \\
\midrule
\textit{Mean} &  &  & 0.972 & 0.993 & 0.693 & 0.908 & +0.021 & +0.215 \\
\bottomrule
\end{tabular}
\end{table}

\subsection{Comparison with dedicated safety classifiers}
\label{sec:classifier_comparison_results}

We benchmark four publicly available safety classifiers on the
same evaluation set (Table~\ref{tab:classifiers},
Appendix~\ref{app:classifier_comparison}). Llama Guard~3 (8B)
and WildGuard (7B) outperform our probe (AUROC~$0.994$ /
TPR~$0.932$ and AUROC $0.998$ / TPR $0.996$). ShieldGemma (9B)
and Latent Guard (host model) score lower: ShieldGemma at
AUROC~$0.939$ / TPR~$0.304$, Latent Guard at AUROC~$0.959$ / TPR~$0.576$.
ShieldGemma's near-ceiling AUROC at TPR $0.304$ illustrates the
pattern from Section~\ref{sec:main_results} at
deployment-relevant scale: roughly $70\%$ of harmful inputs
evade detection at operationally tolerable false-alarm rates,
with the gap most severe on hard-benign XSTest (TPR
$0.145$--$0.305$). Latency and per-source breakdown are in
Appendix~\ref{app:classifier_comparison}.

\section{Discussion}
\label{sec:discussion}

\paragraph{The geometry of the recovered harm direction.}
The harm direction recovered by mean-difference fitting and
Soft-AUC optimisation is concentrated rather than spread across
many directions: refitting in the subspace orthogonal to the
recovered direction at the operating layer drops AUROC
substantially under $\mathrm{mp/raw}$ extraction (mean $0.598$
across the four instruction-tuned models tested) and partially
under $\mathrm{lt/chat}$ extraction (mean $0.780$;
Appendix~\ref{app:self_projection}).
The harm signal is not encoded uniformly along a wide hyperplane
of equally viable scoring axes.
Different extraction protocols, however, recover concentrations
that are nearly orthogonal in the same residual-stream space.
At $\Lp{lt/chat}$, projecting $\wmdp{lt/chat}$ out of
activations under either max-pool extraction
($\mathrm{mp/raw}$ or $\mathrm{mp/chat}$) leaves AUROC
essentially unchanged (difference within $\pm 0.020$); the
mean-difference vector under $\mathrm{mp/raw}$ retains
$\geq 99\%$ of its norm, and under $\mathrm{mp/chat}$ retains
$89$--$98\%$ (Appendix~\ref{app:cross_protocol_projection}).
Under $\mathrm{lt/chat}$ extraction at the same layer,
$\wmdp{lt/chat}$ accounts for nearly all of the
mean-difference vector, and the same projection drops AUROC by
$0.19$--$0.40$.
The recovered harm direction is therefore concentrated within
each protocol but protocol-dependent across them: detection-level
evidence identifies a direction along which a specific extraction
of activations separates harmful from benign prompts, not the
feature the model uses computationally.
Probing methods that recover a single direction and treat it as 
the harm representation report a protocol-specific concentration; 
they do not isolate a unique underlying feature.

\paragraph{Recognition and refusal as separable signals.}
\citet{zhao2025llms} established through causal steering that
harmfulness and refusal are encoded separately in instruction-tuned
models, with steering along each producing dissociable behavioural
effects.
Our results corroborate this finding through different
methodology and at substantially greater breadth.
AUROC differences between abliterated variants and their
instruction-tuned counterparts are within $\pm 0.003$ across four
families: removing the refusal direction from model weights
leaves an input-side recognition signal intact at content-token
positions.
Two pooling choices applied to the same chat-templated 
activations also recover non-collinear directions: 
$\woptp{mp/chat}$ and $\wmdp{lt/chat}$ are at unsigned 
angle $72.6^\circ \pm 6.7^\circ$ at $\Lp{mp/chat}$ and 
$74.9^\circ \pm 9.3^\circ$ at $\Lp{lt/chat}$, consistent 
across both layer choices (Section~\ref{sec:geometry}).
The cross-protocol projection result above strengthens the 
dissociation: $\wmdp{lt/chat}$ has near-zero presence in the 
mean-difference direction recovered by max-pool extraction at 
the same layer, indicating that aggregating activations over 
content tokens versus reading them at the post-instruction 
position produces different mean-difference directions, each 
concentrated along essentially one axis under its own protocol.
Whether the protocols recover different views of one underlying
feature or distinct concentrations of separate features is not
determined by these measurements.

\paragraph{Alignment can rotate harm geometry: the Gemma-3 case.}
Three of four families show base-to-instruct angular shifts of
$17^\circ$ to $39^\circ$ with cross-variant AUROC transfer within
$0.010$ of own-direction performance.
Gemma-3 deviates: alignment rotates the harm direction by
$73^\circ$, base-to-instruct AUROC drops by $0.057$, and 
TPR@1\%FPR collapses from $0.751$ to $0.175$.
Each Gemma-3 variant's own direction recovers AUROC above 
$0.969$, so the harm signal is recoverable; the specific 
decoding direction has moved.
A reading consistent with both this observation and the
within-protocol concentration finding is that alignment in some
architectures rotates the concentrated harm direction in 
residual space without disrupting the linear separability of 
harmful inputs.
What distinguishes Gemma-3 from the other three families at the
mechanism level is open.

\paragraph{AUROC and the deployment regime.}
Among supervised strategies that all exceed AUROC $0.96$,
TPR@1\%FPR varies by more than ten times the AUROC gap, and
this is not a small-probe artefact: ShieldGemma at $9$B 
parameters reaches AUROC $0.939$ with TPR@1\%FPR of $0.304$, 
with the gap most severe on hard-benign XSTest 
(TPR $0.145$--$0.305$). In the near-ceiling regime in which 
safety-adjacent detection lives, AUROC alone is the wrong 
evaluation summary; safety-adjacent detection should report at 
least one operating-point metric alongside AUROC, with the FPR 
as low as available benign data supports and the benign 
distribution stated explicitly.

\paragraph{Limitations.}
\label{sec:limitations}
The evaluation is single-turn, English, and clean.
Whether harm directions remain detectable under adversarial
suffixes~\citep{zou2023universal}, multi-turn escalation,
stylistically diverse harm, or cross-lingual substitution is
untested, and given prior evidence that activation-based methods
can be fragile to adversarial input~\citep{zhu2023autodan}, this
is the most likely source of genuine fragility in the linear
picture.
A second limit follows from the geometry the Discussion has
emphasised: our procedure recovers a protocol-specific
concentrated direction rather than the feature the model uses
computationally.
Identifying the model's actual harm-recognition mechanism
requires causal steering of the kind reported
in~\citet{zhao2025llms} or decomposition through sparse
autoencoders that constrain the basis to align with model-level
features.

\paragraph{Broader impact.}
Activation-based detection adds $\approx4$ms to an existing forward pass, making it practical as a complement to output-side safety classifiers. AUROC alone can obscure low-FPR behaviour, so reporting operating-point metrics remains important. Activation probes can be evaded by adversarial inputs not evaluated here (see Limitations). The abliteration results do not introduce new attack surface, as the underlying procedure is already public~\citep{arditi2024refusal}.

\section{Conclusion}
\label{sec:conclusion}

Harmful intent is recoverable from residual-stream activations as 
a stable linear direction across 12 models spanning four 
architectural families, three alignment variants including 
abliterated models, and three held-out harm benchmarks; 100 
labelled examples per class suffice to fit it. Within-family 
scale stability holds up to 9B in Qwen3.5.
The recovered direction is concentrated within each extraction
protocol but protocol-dependent across them: at the same 
residual-stream layer, the mean-difference direction under 
last-token chat-templated extraction and the mean-difference 
directions under max-pool extraction (raw or chat-templated) are 
near-orthogonal, and projecting the former out leaves the latter 
essentially intact.
Probing identifies a direction along which a specific extraction
of activations separates harmful from benign prompts; it does
not isolate a unique computational feature.
Methodologically, the near-ceiling AUROC regime in which
safety-adjacent detection operates can mask substantial
operating-point variation, demonstrated by a deployed $9$B
classifier reaching AUROC $0.94$ at TPR@1\%FPR of $0.30$;
operating-point metrics alongside AUROC should be reported as a
default in safety-adjacent detection evaluation.
Whether the linear picture survives adversarial pressure,
multi-turn contexts, and languages other than English is the
natural next question.

\bibliography{references}
\bibliographystyle{plainnat}
\newpage
\appendix

\section*{Appendix}

The appendix is organised in the order in which sections are
referenced in the main text.

\begin{itemize}
    \item Section~\ref{app:datasets} reports per-source dataset details
referenced in Section~\ref{sec:setup}.
    \item Section~\ref{app:softauc} provides Soft-AUC optimisation details
referenced in Section~\ref{sec:directions}.
    \item Section~\ref{app:full_results} reports full per-model AUROC and
TPR@1\%FPR with bootstrap confidence intervals, referenced
throughout Section~\ref{sec:results}.
    \item Section~\ref{app:layer_profiles} reports per-layer AUROC for the
supervised strategies across all 12 models, including the 
Qwen2.5 valley profile referenced in 
Section~\ref{sec:geometry}.
    \item Section~\ref{app:score_dist} reports score distributions at the
operating layer.
    \item Section~\ref{app:angles} reports pairwise direction angles at
$\Lp{mp/raw}$ for all 12 models.
    \item Section~\ref{app:projection_test} reports the
refit-after-projecting-$\wmdp{lt/chat}$ test referenced in
Section~\ref{sec:geometry}.
    \item Section~\ref{app:self_projection} reports the within-protocol
projection-and-refit test, in which each protocol's own
mean-difference direction is projected out of activations under
that protocol and a fresh direction is refitted from scratch
($\mathrm{mp/raw}$ and $\mathrm{lt/chat}$).
    \item Section~\ref{app:cross_protocol_projection} reports the
cross-protocol projection-and-refit test, in which
$\wmdp{lt/chat}$ is projected out of activations under three
extraction protocols ($\mathrm{lt/chat}$, $\mathrm{mp/chat}$,
$\mathrm{mp/raw}$).
    \item Section~\ref{app:ood} reports disaggregated out-of-distribution
results.
    \item Section~\ref{app:cross_variant} reports the full $3 \times 3$
cross-variant transfer matrices.
    \item Section~\ref{app:scale} reports the scale-stability extension to
Qwen3.5 at 2B, 4B, and 9B parameters.
    \item Section~\ref{app:sample_efficiency} reports sample efficiency
curves.
    \item Section~\ref{app:classifier_comparison} reports the safety 
classifier benchmark (headline AUROC and TPR@1\%FPR, latency, 
and per-source breakdown).
    \item Section~\ref{app:linearity} discusses the non-uniqueness of
fitted linear directions referenced in
Section~\ref{sec:discussion}.
\end{itemize}
\section{Dataset details}
\label{app:datasets}

We report per-source counts and split assignments
(Table~\ref{tab:datasets_full}). The fit and validation sets 
draw exclusively from AdvBench (harmful) and Alpaca-Cleaned 
(benign). HarmBench, JailbreakBench, and XSTest are held out 
entirely from fitting and used only at evaluation. The Alpaca 
evaluation split is similarly held out from the fit and 
validation splits.

\paragraph{Prompt normalisation rationale.}
Without terminal-punctuation stripping, AdvBench prompts (which
uniformly end in a full stop) are trivially separable from Alpaca
prompts (which typically do not). A one-feature classifier on 
terminal punctuation achieves AUROC $> 0.99$ on the unnormalised 
data; this confounder is removed before any geometry analysis.

\begin{table}[ht]
\centering\small
\caption{Dataset sources, splits, and normalisation outcomes.
  \%~punct: fraction of prompts with terminal punctuation removed
  by normalisation.}
\label{tab:datasets_full}
\begin{tabular}{llrrrr}
\toprule
Source & Pillar & $n_{\mathrm{fit}}$ & $n_{\mathrm{val}}$ & $n_{\mathrm{eval}}$ & \% punct \\
\midrule
AdvBench               & I (canonical harm)        & 100 & 50 & 370 & 100\% \\
HarmBench              & I (canonical harm)        &   0 &  0 & 200 & 0\% \\
JailbreakBench         & II (adversarial)          &   0 &  0 & 100 & 0\% \\
Alpaca-Cleaned         & Benign                    & 100 & 50 & 500 & 0.1\% \\
XSTest (safe subset)   & III (hard benign)         &   0 &  0 & 250 & 0\% \\
\bottomrule
\end{tabular}
\end{table}

\paragraph{Model checkpoints.}
Table~\ref{tab:model_ids} lists the exact HuggingFace identifiers
for the 12 models in our main analysis. Abliterated variants are 
community releases of the procedure of~\citet{arditi2024refusal} 
applied to the corresponding instruct checkpoints. Each 
abliterated variant inherits its tokenizer from the corresponding 
instruct checkpoint and is loaded with the same 
\texttt{transformers} pipeline. We did not retrain or modify 
these checkpoints; activation extraction uses the same 
forward-hook procedure as for base and instruct variants.

\begin{table}[ht]
\centering\small
\caption{HuggingFace model identifiers for the 12 models evaluated. All loaded in \texttt{bfloat16} on a
single NVIDIA RTX~3070 Mobile (8~GB VRAM). $L$: transformer
layers. $D$: residual-stream dimension.}
\label{tab:model_ids}
\begin{tabular}{lllcc}
\toprule
Family & Variant & HuggingFace ID & $L$ & $D$ \\
\midrule
Qwen2.5 & base        & \texttt{Qwen/Qwen2.5-0.5B} & 24 & 896\\
        & instruct    & \texttt{Qwen/Qwen2.5-0.5B-Instruct} & 24 & 896\\
        & abliterated & \texttt{huihui-ai/Qwen2.5-0.5B-Instruct-abliterated} & 24 & 896\\
\midrule
Qwen3.5 & base        & \texttt{Qwen/Qwen3.5-0.8B-Base} & 24 & 1024\\
        & instruct    & \texttt{Qwen/Qwen3.5-0.8B} & 24 & 1024\\
        & abliterated & \texttt{huihui-ai/Huihui-Qwen3.5-0.8B-abliterated} & 24 & 1024\\
\midrule
Llama-3.2 & base        & \texttt{meta-llama/Llama-3.2-1B} & 16 & 2048\\
          & instruct    & \texttt{meta-llama/Llama-3.2-1B-Instruct} & 16 & 2048\\
          & abliterated & \texttt{huihui-ai/Llama-3.2-1B-Instruct-abliterated} & 16 & 2048\\
\midrule
Gemma-3 & base        & \texttt{google/gemma-3-1b-pt} & 26 & 1152\\
        & instruct    & \texttt{google/gemma-3-1b-it} & 26 & 1152\\
        & abliterated & \texttt{huihui-ai/gemma-3-1b-it-abliterated} & 26 & 1152\\
\bottomrule
\end{tabular}
\end{table}

\paragraph{Asset licenses and access.}
All datasets and models used in this paper are publicly 
released. Datasets: AdvBench~\citep{zou2023universal} (MIT), 
HarmBench~\citep{mazeika2024harmbench} (MIT), 
JailbreakBench~\citep{chao2024jailbreakbench} (MIT), 
Alpaca-Cleaned~\citep{taori2023alpaca} (CC BY-NC 4.0), 
XSTest~\citep{rottger2024xstest} (CC-BY-4.0). Model checkpoints 
are accessed via HuggingFace under the licenses specified by 
each publisher. Several checkpoints require accepting the 
publisher's terms of use before download: Llama-3.2-1B and 
Llama-3.2-1B-Instruct (Llama~3.2 Community License), Gemma-3-1B 
variants (Gemma Terms of Use), and Llama Guard~3 (Llama~3.1 
Community License). Qwen2.5, Qwen3.5 and WildGuard are released under Apache~2.0 with no gating. 
Latent Guard~\citep{zhao2025llms} does not cite any license.
Abliterated variants released by \texttt{huihui-ai} inherit the license of 
the corresponding base checkpoint. Users seeking to reproduce 
this work need to register on HuggingFace and accept the 
relevant terms before downloading gated assets; the released 
code provides automated handling of the download once access 
is granted.

\section{Soft-AUC optimisation details}
\label{app:softauc}

The Soft-AUC objective (Equation~\ref{eq:soft_auc}) is optimised
via Riemannian gradient ascent on the unit sphere $S^{D-1}$. At 
each step, the Euclidean gradient $\nabla \hat{U}$ is projected 
onto the tangent space of the sphere at $\mathbf{w}$ to obtain 
the Riemannian gradient:
\begin{equation}
  \nabla_R \hat{U}
  = \nabla \hat{U}
    - (\nabla \hat{U} \cdot \mathbf{w})\, \mathbf{w}.
  \label{eq:riem_grad}
\end{equation}
The objective is computed without a temperature scale 
(equivalent to $\tau = 1$ in standard formulations); the 
optimisation is invariant to global scaling of activations on 
the unit sphere. The update rule is $\mathbf{w} \leftarrow 
\mathrm{unit}(\mathbf{w} + \eta \nabla_R \hat{U})$, where 
$\mathrm{unit}(\cdot)$ denotes $\ell_2$ normalisation and 
$\eta = 0.05$ is the step size. The optimisation is warm-started 
from the protocol's mean-difference direction $\wmdp{P/F}$
(Equation~\ref{eq:mean_diff}) when fitting $\woptp{P/F}$ for any 
protocol. Early stopping is triggered when the Riemannian 
gradient norm satisfies $\|\nabla_R \hat{U}\| < 10^{-5}$ for 20 
consecutive steps, up to a maximum of 300 steps. If a NaN 
gradient is encountered at any step, the optimiser reverts to 
the previous valid iterate and terminates.

\section{Full per-model results}
\label{app:full_results}
\begin{figure}[ht]
\centering
\includegraphics[width=\textwidth]{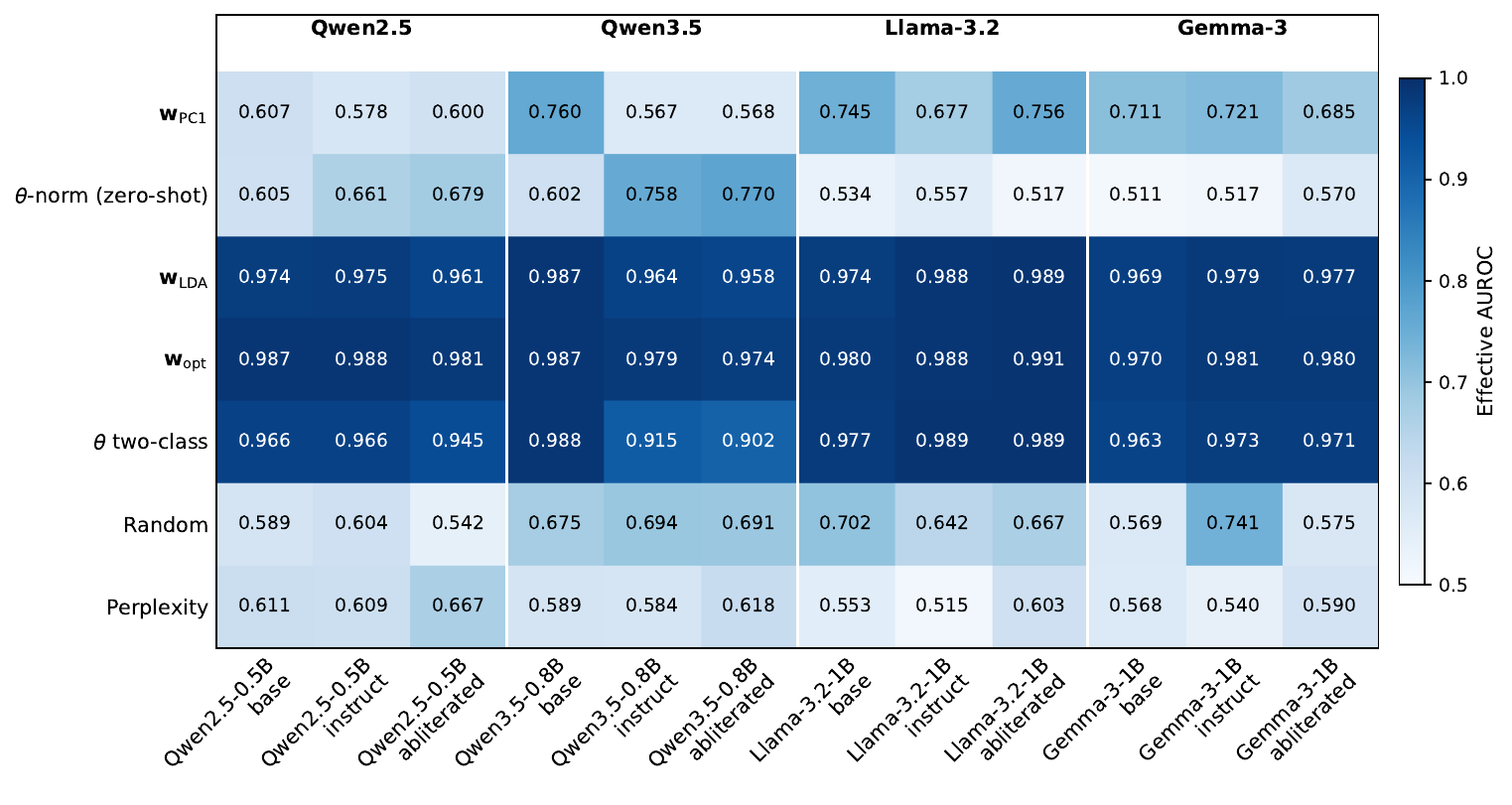}
\caption{Effective AUROC across all 12 models and seven 
detection strategies under $\mathrm{mp/raw}$ extraction at 
$\Lp{mp/raw}$. Each cell is one (strategy, model) pair. 
Supervised strategies ($\wmd$, $\wopt$, $\theta$ two-class) 
exceed AUROC $0.90$ on every model; unsupervised strategies 
(PC1, $\theta$-norm) and perplexity vary substantially across 
models. Numerical values with bootstrap 95\% CIs in 
Table~\ref{tab:full_a}.}
\label{fig:auroc_heatmap}
\end{figure}

\begin{figure}[ht]
\centering
\includegraphics[width=\textwidth]{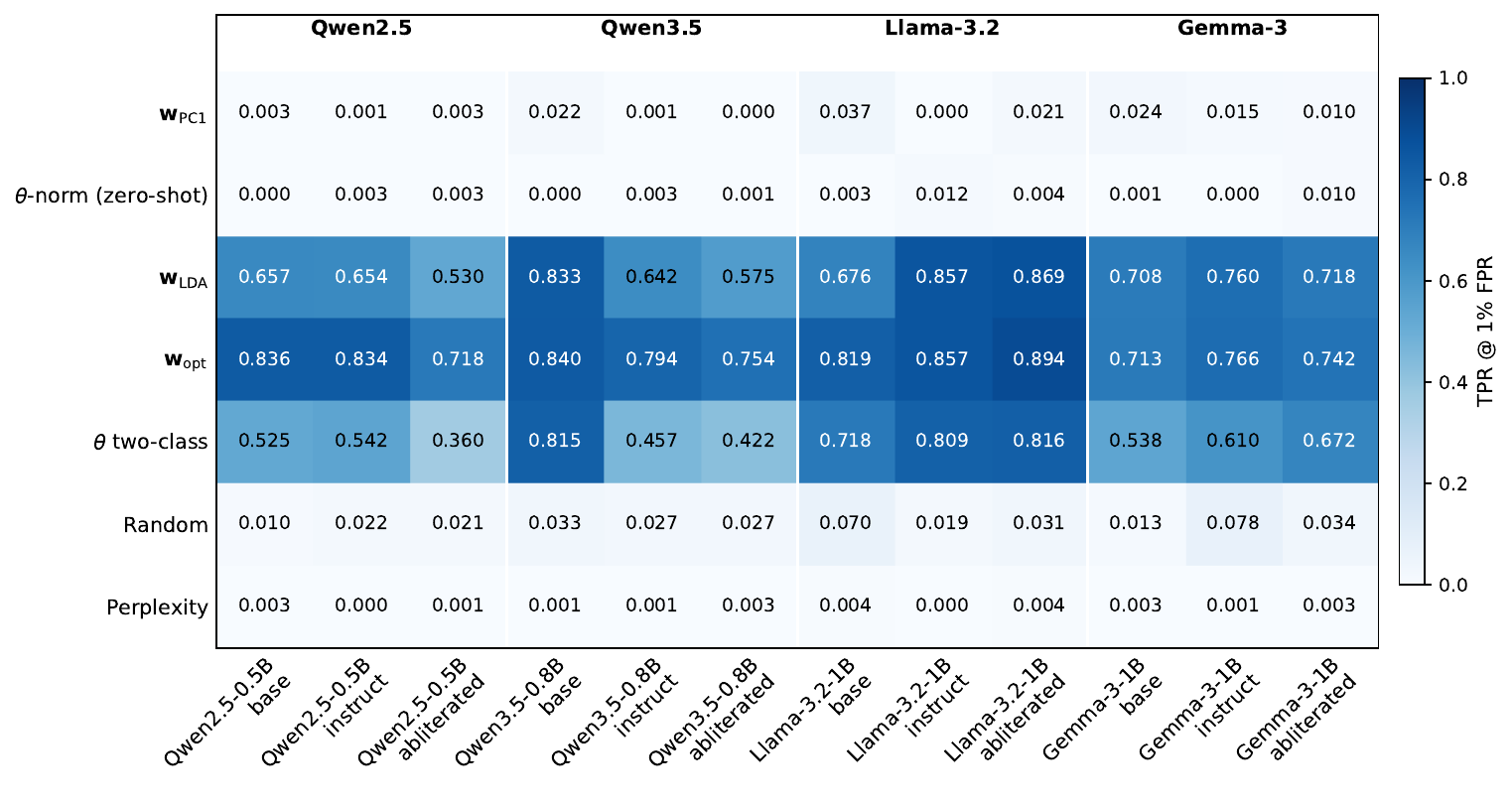}
\caption{TPR@1\%FPR across all 12 models and seven detection 
strategies under $\mathrm{mp/raw}$ extraction at $\Lp{mp/raw}$. 
The contrast with Figure~\ref{fig:auroc_heatmap} illustrates 
the AUROC--TPR divergence discussed in 
Section~\ref{sec:main_results}: zero-shot baselines reach AUROC 
$0.50$--$0.78$ but TPR below $0.05$ on every model; supervised 
strategies sustain TPR $0.36$--$0.89$. Numerical values with 
bootstrap 95\% CIs in Table~\ref{tab:full_b}.}
\label{fig:tpr_heatmap}
\end{figure}

Figures~\ref{fig:auroc_heatmap} and~\ref{fig:tpr_heatmap} 
visualise the per-model effective AUROC and TPR@1\%FPR for all 
six direction strategies plus baselines. The supervised 
strategies ($\wmd$, $\wopt$, $\theta$ two-class) recover 
near-ceiling AUROC and substantial TPR across all 12 models; 
zero-shot strategies and the perplexity baseline remain near 
zero in TPR even where their AUROC is non-trivial, illustrating 
the AUROC-TPR divergence discussed in 
Section~\ref{sec:main_results}. Tables~\ref{tab:full_a} 
and~\ref{tab:full_b} report the same values numerically with 
stratified bootstrap 95\% confidence intervals.

Tables~\ref{tab:full_a} and~\ref{tab:full_b} report effective 
AUROC and TPR@1\%FPR for all six direction strategies plus 
baselines, evaluated against the full benign set (Alpaca + 
XSTest) at the validation-selected layer for each model under 
$\mathrm{mp/raw}$ extraction. Stratified bootstrap 95\% 
confidence intervals (1{,}000 resamples, stratified by 
evaluation source) appear in brackets. Bold marks the highest 
AUROC within each model block.

\paragraph{Per-source variation in zero-shot performance.}
While the model-averaged AUROC for $\theta$-normative is below 
$0.78$ across the 12 models, the per-source breakdown 
(Tables~\ref{tab:full_a} and~\ref{tab:full_b}) shows substantial 
variation. On Qwen3.5-Instruct and Qwen3.5-abliterated against 
XSTest, $\theta$-normative reaches AUROC $0.957$ and $0.971$ 
respectively. The zero-shot baseline is not uniformly poor; 
specific (model, dataset) cells exceed AUROC $0.95$. The 
averaged comparison still favours supervised methods 
substantially in TPR@1\%FPR (zero-shot baselines remain below 
$0.05$ TPR on every model).

\begin{table}[ht]
\centering\footnotesize
\setlength{\tabcolsep}{3pt}
\caption{Full per-model results, Part~1: Qwen2.5 and Qwen3.5. All metrics evaluated against the full benign set (Alpaca\,+\,XSTest) at the validation-selected layer. Bold marks the highest AUROC median within each model. Stratified bootstrap 95\% CIs (1{,}000 resamples) in brackets.}
\label{tab:full_a}
\begin{tabular}{llrll}
\toprule
Model & Strategy & Layer & AUROC [95\% CI] & TPR@1\%FPR [95\% CI] \\
\midrule
  \multirow{7}{*}{\shortstack[l]{Qwen2.5\\(base)}} & $\wmd$ & 22 & 0.974 [0.966, 0.980] & 0.657 [0.495, 0.763] \\
   & $\wopt$ & 22 & \textbf{0.987} [0.982, 0.991] & 0.836 [0.758, 0.887] \\
   & $\wpca$ & 22 & 0.607 [0.582, 0.634] & 0.003 [0.000, 0.027] \\
   & $\theta$-norm & 22 & 0.605 [0.578, 0.632] & 0.000 [0.000, 0.000] \\
   & $\theta$ two-class & 22 & 0.966 [0.958, 0.974] & 0.525 [0.304, 0.701] \\
   & Random & 22 & 0.589 [0.562, 0.615] & 0.010 [0.003, 0.031] \\
   & Perplexity & --- & 0.611 [0.587, 0.635] & 0.003 [0.000, 0.009] \\
\midrule
  \multirow{7}{*}{\shortstack[l]{Qwen2.5\\(instruct)}} & $\wmd$ & 22 & 0.975 [0.968, 0.981] & 0.654 [0.545, 0.737] \\
   & $\wopt$ & 22 & \textbf{0.988} [0.983, 0.992] & 0.834 [0.707, 0.885] \\
   & $\wpca$ & 22 & 0.578 [0.551, 0.605] & 0.001 [0.000, 0.021] \\
   & $\theta$-norm & 22 & 0.661 [0.634, 0.686] & 0.003 [0.000, 0.009] \\
   & $\theta$ two-class & 22 & 0.966 [0.957, 0.974] & 0.542 [0.334, 0.703] \\
   & Random & 22 & 0.604 [0.576, 0.631] & 0.022 [0.007, 0.046] \\
   & Perplexity & --- & 0.609 [0.582, 0.636] & 0.000 [0.000, 0.013] \\
\midrule
  \multirow{7}{*}{\shortstack[l]{Qwen2.5\\(abliterated)}} & $\wmd$ & 22 & 0.961 [0.952, 0.969] & 0.530 [0.330, 0.624] \\
   & $\wopt$ & 22 & \textbf{0.981} [0.975, 0.986] & 0.718 [0.455, 0.840] \\
   & $\wpca$ & 22 & 0.600 [0.572, 0.627] & 0.003 [0.000, 0.024] \\
   & $\theta$-norm & 22 & 0.679 [0.653, 0.704] & 0.003 [0.000, 0.018] \\
   & $\theta$ two-class & 22 & 0.945 [0.934, 0.956] & 0.360 [0.191, 0.530] \\
   & Random & 22 & 0.542 [0.515, 0.569] & 0.021 [0.007, 0.036] \\
   & Perplexity & --- & 0.667 [0.641, 0.693] & 0.001 [0.000, 0.018] \\
\midrule
  \multirow{7}{*}{\shortstack[l]{Qwen3.5\\(base)}} & $\wmd$ & 10 & 0.987 [0.982, 0.990] & 0.833 [0.761, 0.881] \\
   & $\wopt$ & 10 & 0.987 [0.983, 0.991] & 0.840 [0.772, 0.884] \\
   & $\wpca$ & 10 & 0.760 [0.737, 0.783] & 0.022 [0.007, 0.039] \\
   & $\theta$-norm & 10 & 0.602 [0.574, 0.630] & 0.000 [0.000, 0.013] \\
   & $\theta$ two-class & 10 & \textbf{0.988} [0.984, 0.992] & 0.815 [0.745, 0.869] \\
   & Random & 10 & 0.675 [0.650, 0.703] & 0.033 [0.016, 0.057] \\
   & Perplexity & --- & 0.589 [0.562, 0.615] & 0.001 [0.000, 0.004] \\
\midrule
  \multirow{7}{*}{\shortstack[l]{Qwen3.5\\(instruct)}} & $\wmd$ & 22 & 0.964 [0.955, 0.972] & 0.642 [0.558, 0.701] \\
   & $\wopt$ & 22 & \textbf{0.979} [0.972, 0.984] & 0.794 [0.743, 0.827] \\
   & $\wpca$ & 22 & 0.567 [0.539, 0.594] & 0.001 [0.000, 0.018] \\
   & $\theta$-norm & 22 & 0.758 [0.735, 0.781] & 0.003 [0.000, 0.013] \\
   & $\theta$ two-class & 22 & 0.915 [0.902, 0.927] & 0.457 [0.366, 0.533] \\
   & Random & 22 & 0.694 [0.666, 0.719] & 0.027 [0.006, 0.054] \\
   & Perplexity & --- & 0.584 [0.556, 0.609] & 0.001 [0.000, 0.006] \\
\midrule
  \multirow{7}{*}{\shortstack[l]{Qwen3.5\\(abliterated)}} & $\wmd$ & 22 & 0.958 [0.949, 0.966] & 0.575 [0.524, 0.643] \\
   & $\wopt$ & 22 & \textbf{0.974} [0.967, 0.980] & 0.754 [0.663, 0.791] \\
   & $\wpca$ & 22 & 0.568 [0.540, 0.595] & 0.000 [0.000, 0.018] \\
   & $\theta$-norm & 22 & 0.770 [0.748, 0.793] & 0.001 [0.000, 0.015] \\
   & $\theta$ two-class & 22 & 0.902 [0.888, 0.915] & 0.422 [0.304, 0.493] \\
   & Random & 22 & 0.691 [0.663, 0.717] & 0.027 [0.003, 0.051] \\
   & Perplexity & --- & 0.618 [0.592, 0.642] & 0.003 [0.000, 0.007] \\
\bottomrule
\end{tabular}
\end{table}

\begin{table}[htp]
\centering\footnotesize
\setlength{\tabcolsep}{3pt}
\caption{Full per-model results, Part~2: Llama-3.2 and Gemma-3. All metrics evaluated against the full benign set (Alpaca\,+\,XSTest) at the validation-selected layer. Bold marks the highest AUROC median within each model. Stratified bootstrap 95\% CIs (1{,}000 resamples) in brackets.}
\label{tab:full_b}
\begin{tabular}{llrll}
\toprule
Model & Strategy & Layer & AUROC [95\% CI] & TPR@1\%FPR [95\% CI] \\
\midrule
  \multirow{7}{*}{\shortstack[l]{Llama-3.2\\(base)}} & $\wmd$ & 6 & 0.974 [0.966, 0.981] & 0.676 [0.539, 0.782] \\
   & $\wopt$ & 6 & \textbf{0.980} [0.973, 0.985] & 0.819 [0.658, 0.860] \\
   & $\wpca$ & 6 & 0.745 [0.718, 0.769] & 0.037 [0.009, 0.061] \\
   & $\theta$-norm & 6 & 0.534 [0.508, 0.562] & 0.003 [0.000, 0.010] \\
   & $\theta$ two-class & 6 & 0.977 [0.970, 0.983] & 0.718 [0.461, 0.851] \\
   & Random & 6 & 0.702 [0.674, 0.728] & 0.070 [0.042, 0.115] \\
   & Perplexity & --- & 0.553 [0.524, 0.585] & 0.004 [0.000, 0.010] \\
\midrule
  \multirow{7}{*}{\shortstack[l]{Llama-3.2\\(instruct)}} & $\wmd$ & 5 & 0.988 [0.982, 0.992] & 0.857 [0.766, 0.906] \\
   & $\wopt$ & 5 & 0.988 [0.982, 0.992] & 0.857 [0.775, 0.909] \\
   & $\wpca$ & 5 & 0.677 [0.649, 0.704] & 0.000 [0.000, 0.006] \\
   & $\theta$-norm & 5 & 0.557 [0.530, 0.585] & 0.012 [0.003, 0.028] \\
   & $\theta$ two-class & 5 & \textbf{0.989} [0.984, 0.992] & 0.809 [0.733, 0.899] \\
   & Random & 5 & 0.642 [0.611, 0.670] & 0.019 [0.006, 0.042] \\
   & Perplexity & --- & 0.515 [0.501, 0.542] & 0.000 [0.000, 0.007] \\
\midrule
  \multirow{7}{*}{\shortstack[l]{Llama-3.2\\(abliterated)}} & $\wmd$ & 5 & 0.989 [0.985, 0.992] & 0.869 [0.730, 0.918] \\
   & $\wopt$ & 5 & \textbf{0.991} [0.987, 0.994] & 0.894 [0.743, 0.928] \\
   & $\wpca$ & 5 & 0.756 [0.731, 0.779] & 0.021 [0.000, 0.051] \\
   & $\theta$-norm & 5 & 0.517 [0.501, 0.545] & 0.004 [0.000, 0.010] \\
   & $\theta$ two-class & 5 & 0.989 [0.984, 0.992] & 0.816 [0.675, 0.903] \\
   & Random & 5 & 0.667 [0.640, 0.692] & 0.031 [0.018, 0.054] \\
   & Perplexity & --- & 0.603 [0.575, 0.629] & 0.004 [0.000, 0.015] \\
\midrule
  \multirow{7}{*}{\shortstack[l]{Gemma-3\\(base)}} & $\wmd$ & 19 & 0.969 [0.960, 0.976] & 0.708 [0.612, 0.772] \\
   & $\wopt$ & 19 & \textbf{0.970} [0.961, 0.977] & 0.713 [0.633, 0.775] \\
   & $\wpca$ & 19 & 0.711 [0.684, 0.738] & 0.024 [0.007, 0.070] \\
   & $\theta$-norm & 19 & 0.511 [0.501, 0.536] & 0.001 [0.000, 0.009] \\
   & $\theta$ two-class & 19 & 0.963 [0.953, 0.971] & 0.538 [0.451, 0.709] \\
   & Random & 19 & 0.569 [0.539, 0.597] & 0.013 [0.004, 0.030] \\
   & Perplexity & --- & 0.568 [0.537, 0.599] & 0.003 [0.000, 0.012] \\
\midrule
  \multirow{7}{*}{\shortstack[l]{Gemma-3\\(instruct)}} & $\wmd$ & 19 & 0.979 [0.973, 0.985] & 0.760 [0.655, 0.813] \\
   & $\wopt$ & 19 & \textbf{0.981} [0.974, 0.986] & 0.766 [0.657, 0.822] \\
   & $\wpca$ & 19 & 0.721 [0.695, 0.747] & 0.015 [0.000, 0.070] \\
   & $\theta$-norm & 19 & 0.517 [0.501, 0.543] & 0.000 [0.000, 0.009] \\
   & $\theta$ two-class & 19 & 0.973 [0.965, 0.980] & 0.610 [0.491, 0.767] \\
   & Random & 19 & 0.741 [0.715, 0.768] & 0.078 [0.042, 0.130] \\
   & Perplexity & --- & 0.540 [0.511, 0.568] & 0.001 [0.000, 0.006] \\
\midrule
  \multirow{7}{*}{\shortstack[l]{Gemma-3\\(abliterated)}} & $\wmd$ & 19 & 0.977 [0.970, 0.983] & 0.718 [0.664, 0.785] \\
   & $\wopt$ & 19 & \textbf{0.980} [0.973, 0.985] & 0.742 [0.678, 0.806] \\
   & $\wpca$ & 19 & 0.685 [0.657, 0.713] & 0.010 [0.000, 0.037] \\
   & $\theta$-norm & 19 & 0.570 [0.541, 0.598] & 0.010 [0.000, 0.033] \\
   & $\theta$ two-class & 19 & 0.971 [0.963, 0.978] & 0.672 [0.503, 0.763] \\
   & Random & 19 & 0.575 [0.544, 0.606] & 0.034 [0.012, 0.070] \\
   & Perplexity & --- & 0.590 [0.563, 0.617] & 0.003 [0.000, 0.009] \\
\bottomrule
\end{tabular}
\end{table}

\section{Per-layer AUROC profiles}
\label{app:layer_profiles}

\begin{figure}[ht]
\centering
\includegraphics[width=\textwidth]{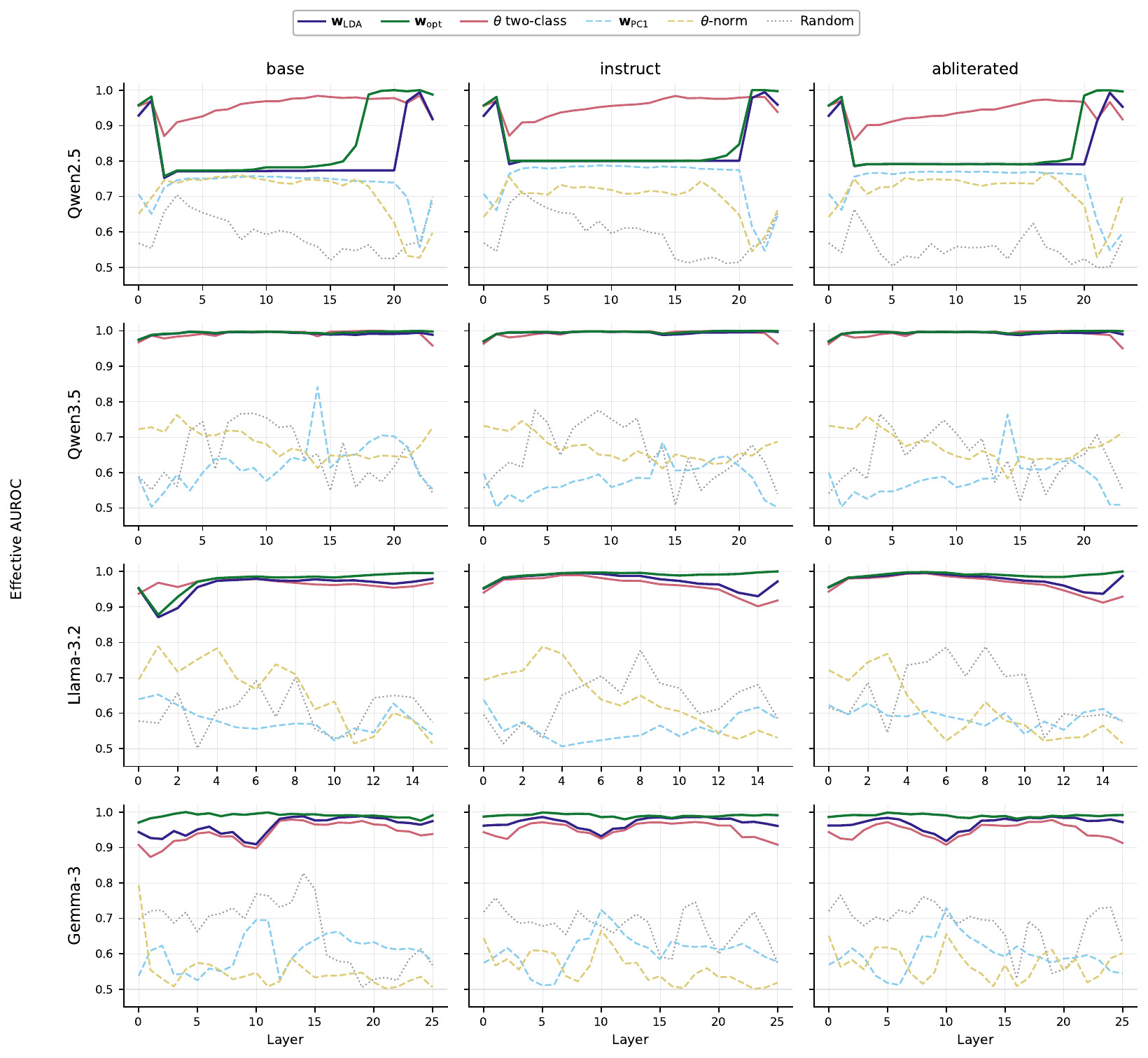}
\caption{Per-layer effective AUROC for the supervised 
  strategies across all 12 models, organised by family (rows) 
  and alignment variant (columns), under $\mathrm{mp/raw}$ 
  extraction. Three families (Qwen3.5, Llama-3.2, Gemma-3) show 
  flat profiles: $\wmdp{mp/raw}$ and $\woptp{mp/raw}$ exceed 
  AUROC $0.92$ from early layers onward. Qwen2.5 shows a valley 
  profile: projection-based detection collapses in middle 
  layers but the $\theta$~two-class angular strategy sustains 
  AUROC above $0.95$ throughout. Each point is an independent 
  fit and evaluation at that layer. Zero-shot baselines (PC1, 
  $\theta$-normative, random, perplexity) are below the 
  supervised curves at every layer.}
\label{fig:layers}
\end{figure}

Figure~\ref{fig:layers} reports per-layer effective AUROC for 
the three supervised strategies ($\wmdp{mp/raw}$, 
$\woptp{mp/raw}$, $\theta$ two-class) across all 12 models 
under $\mathrm{mp/raw}$ extraction. At each layer the strategies 
are independently fitted on the fit set extracted at that layer 
and evaluated on the evaluation set. Layer selection (the 
$\Lp{mp/raw}$ used for headline metrics) follows the procedure 
described in Section~\ref{sec:scoring}.

\paragraph{Two qualitative patterns.}
Three of four families (Qwen3.5, Llama-3.2, Gemma-3) show flat 
profiles for the projection-based strategies: $\wmdp{mp/raw}$ 
and $\woptp{mp/raw}$ exceed AUROC $0.92$ from early layers 
onward and remain near-ceiling thereafter. Qwen2.5 shows a 
valley profile: across layers $2$--$20$ the projection-based 
strategies degrade to AUROC below $0.80$, while 
$\theta$~two-class sustains AUROC above $0.95$ throughout. The 
valley is consistent across the three Qwen2.5 alignment 
variants.

\paragraph{Implications for layer selection.}
The flat profiles in three families mean the validation-selected
layer is not load-bearing for the headline metrics; AUROC at 
the chosen layer is representative of AUROC at adjacent layers. 
The Qwen2.5 valley is the case where layer selection matters 
most for projection-based detection: the validation-selected 
layer ($\Lp{mp/raw} = 22$ for the instruct variant) sits above 
the valley region, recovering AUROC $0.975$. Within the valley, 
projection-based detection collapses while angular detection 
remains near-ceiling. This geometric observation motivates 
the $\theta$~two-class strategy 
(Section~\ref{sec:geometry}).

\section{Score distributions at the operating layer}
\label{app:score_dist}

\begin{figure}[ht]
\centering
\includegraphics[width=\textwidth]{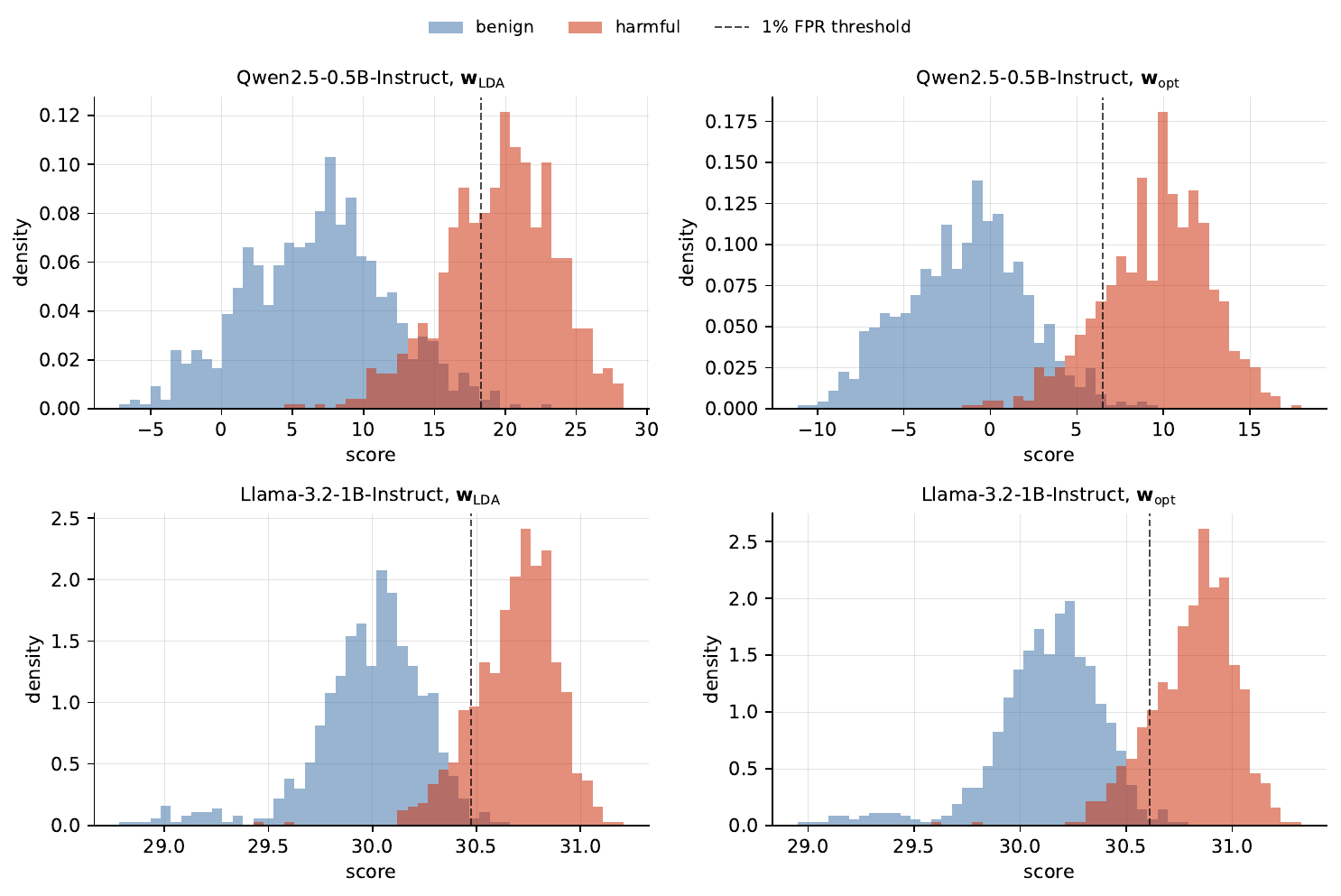}
\caption{Score distributions at the validation-selected layer 
  $\Lp{mp/raw}$ for Qwen2.5-0.5B-Instruct and 
  Llama-3.2-1B-Instruct. Dashed vertical line: 1\%~FPR 
  threshold. Qwen2.5 at layer 22: AUROC 0.975 / TPR 0.654 
  ($\wmdp{mp/raw}$), AUROC 0.988 / TPR 0.834 
  ($\woptp{mp/raw}$). Llama-3.2 at layer 5: both strategies 
  achieve 0.988 AUROC and 0.857 TPR.}
\label{fig:score_dist}
\end{figure}

Figure~\ref{fig:score_dist} shows score distributions at the
validation-selected layer $\Lp{mp/raw}$ for two 
instruction-tuned models, illustrating the AUROC--TPR pattern 
in Section~\ref{sec:main_results}. Qwen2.5-0.5B-Instruct under 
$\wmdp{mp/raw}$ achieves AUROC 0.975 with the right tail of 
the benign distribution overlapping the harmful distribution, 
placing the 1\%~FPR threshold inside the overlap region 
(TPR 0.654). Under $\woptp{mp/raw}$, the overlap shrinks and 
the threshold clears it, lifting TPR to 0.834. 
Llama-3.2-1B-Instruct shows clean separation under both 
strategies, yielding identical TPR (0.857).

\section{Direction geometry}
\label{app:angles}

Table~\ref{tab:angles_full} reports pairwise unsigned angles
between direction vectors at $\Lp{mp/raw}$, aggregated across 
all 12 models. $0^\circ$ corresponds to identical directions; 
$90^\circ$ to orthogonality.

\begin{table}[ht]
\centering\small
\caption{Pairwise unsigned angles between direction vectors at 
  $\Lp{mp/raw}$, aggregated across all 12 main-set models. All 
  directions are fitted under $\mathrm{mp/raw}$ extraction. 
  $n$: number of models contributing to each row.}
\label{tab:angles_full}
\begin{tabular}{lrrrrr}
\toprule
Direction pair & $n$ & Mean & Std & Min & Max \\
\midrule
$\wmdp{mp/raw}$, $\woptp{mp/raw}$               & 12 & $9.1^\circ$  & $8.8^\circ$  & $1.0^\circ$  & $24.7^\circ$ \\
$\wmdp{mp/raw}$, $\theta$ two-class             & 12 & $73.4^\circ$ & $12.3^\circ$ & $57.3^\circ$ & $88.4^\circ$ \\
$\theta$-norm, $\theta$ two-class               & 12 & $20.6^\circ$ & $15.9^\circ$ & $2.4^\circ$  & $42.3^\circ$ \\
$\wmdp{mp/raw}$, $\wpca$ (benign)               & 12 & $81.7^\circ$ & $4.1^\circ$  & $76.2^\circ$ & $87.3^\circ$ \\
$\wpca$ (benign), $\wpca$ (harmful)             & 12 & $64.8^\circ$ & $11.1^\circ$ & $39.1^\circ$ & $77.7^\circ$ \\
$\wmdp{mp/raw}$, random                         & 12 & $88.1^\circ$ & $0.8^\circ$  & $86.4^\circ$ & $89.4^\circ$ \\
\bottomrule
\end{tabular}
\end{table}

\begin{figure}[ht]
\centering
\includegraphics[width=\textwidth]{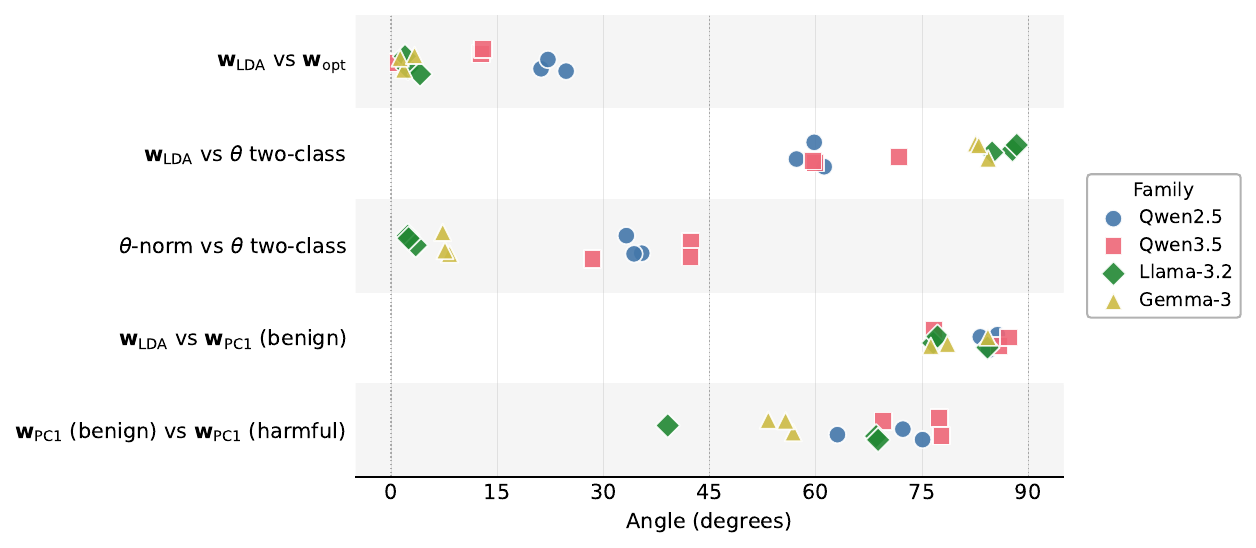}
\caption{Pairwise unsigned angles between direction vectors at
  $\Lp{mp/raw}$ for all 12 models. $0^\circ$: identical 
  directions. $90^\circ$: orthogonal.}
\label{fig:angles_full}
\end{figure}

\section{Refit after projecting out the refusal direction}
\label{app:projection_test}

We test whether harm is linearly recoverable from activations 
after the refusal direction $\wmdp{lt/chat}$ has been removed. 
\textbf{Procedure.} At each instruction-tuned model's 
$\Lp{lt/chat}$, extract activations under $\mathrm{mp/chat}$ 
(max-pool over content-token positions of chat-templated 
input); project $\wmdp{lt/chat}$ out of fit, validation, and 
evaluation activations; refit $\woptp{mp/chat}{}'$ from scratch 
on the projected fit set; evaluate on the projected evaluation 
set. As a comparison condition, we also report the baseline 
$\woptp{mp/chat}$ at $\Lp{lt/chat}$ on unprojected activations 
and the original $\woptp{mp/chat}$ applied directly to 
projected activations (no refit).

Table~\ref{tab:g3_refit} reports AUROC for each condition. 
Projecting $\wmdp{lt/chat}$ out and refitting recovers AUROC 
of $0.949$ on Qwen2.5 (vs.\ $0.929$ unprojected at the same 
layer); on the other three models the refitted AUROC is 
$\geq 0.991$, within bootstrap noise of the unprojected 
baseline.

\begin{table}[ht]
\centering\small
\setlength{\tabcolsep}{4pt}
\caption{Refit of $\woptp{mp/chat}$ at $\Lp{lt/chat}$ for the 
four instruction-tuned models, with the refusal direction 
projected out of activations. \textbf{Protocol.} Activations 
extracted under $\mathrm{mp/chat}$ (max-pool over content-token 
positions of chat-templated input) at $\Lp{lt/chat}$. 
\textbf{Conditions.} \emph{baseline}: $\woptp{mp/chat}$ on 
unprojected activations. \emph{$\wmdp{lt/chat}$ projected}: 
$\woptp{mp/chat}$ applied to activations with $\wmdp{lt/chat}$ 
projected out (no refit). \emph{$\wmdp{lt/chat}$ refit}: 
$\woptp{mp/chat}{}'$ trained from scratch on activations with 
$\wmdp{lt/chat}$ projected out. AUROC with stratified bootstrap 
95\% CIs (1{,}000 resamples).}
\label{tab:g3_refit}
\begin{tabular}{lll}
\toprule
Model & Condition & AUROC [95\% CI] \\
\midrule
  Qwen2.5-0.5B-Instruct & original (baseline) & 0.929 [0.915, 0.941] \\
  Qwen2.5-0.5B-Instruct & original on projected & 0.925 [0.910, 0.938] \\
  Qwen2.5-0.5B-Instruct & refit, $\wmdp{lt/chat}$ removed & 0.949 [0.939, 0.958] \\
\midrule
  Qwen3.5-0.8B & original (baseline) & 0.990 [0.987, 0.993] \\
  Qwen3.5-0.8B & original on projected & 0.988 [0.984, 0.991] \\
  Qwen3.5-0.8B & refit, $\wmdp{lt/chat}$ removed & 0.991 [0.987, 0.993] \\
\midrule
  Llama-3.2-1B-Instruct & original (baseline) & 0.997 [0.995, 0.998] \\
  Llama-3.2-1B-Instruct & original on projected & 0.994 [0.991, 0.996] \\
  Llama-3.2-1B-Instruct & refit, $\wmdp{lt/chat}$ removed & 0.997 [0.995, 0.998] \\
\midrule
  gemma-3-1b-it & original (baseline) & 0.996 [0.994, 0.998] \\
  gemma-3-1b-it & original on projected & 0.997 [0.995, 0.998] \\
  gemma-3-1b-it & refit, $\wmdp{lt/chat}$ removed & 0.997 [0.995, 0.998] \\
\bottomrule
\end{tabular}
\end{table}

\section{Refit after projecting out the harm direction}
\label{app:self_projection}

We test how concentrated the harm signal is along the recovered
direction at each protocol's operating layer by projecting the
direction out of activations and refitting from scratch under
that same protocol. \textbf{Procedure}, applied independently
under two protocols ($\mathrm{mp/raw}$ at $\Lp{mp/raw}$ and 
$\mathrm{lt/chat}$ at $\Lp{lt/chat}$): fit $\wmdp{P/F}$ on the 
fit set; project $\wmdp{P/F}$ out of fit, validation, and 
evaluation activations
($\mathbf{x}' = \mathbf{x} - (\mathbf{x} \cdot \wmdp{P/F})\,
\wmdp{P/F}$);
refit $\woptp{P/F}{}'$ by Soft-AUC optimisation on projected fit
activations; evaluate on projected evaluation activations.
Refitting $\wmdp{P/F}$ on projected activations yields a 
degenerate direction by construction
($\hat{\mu}^{P/F}_\harm - \hat{\mu}^{P/F}_\norm$ is collinear 
with the projected-out direction by definition of $\wmdp{P/F}$);
we therefore report only $\woptp{P/F}{}'$.

Table~\ref{tab:g4_refit} reports AUROC for the baseline 
($\wmdp{P/F}$ and $\woptp{P/F}$ on unprojected activations 
under each protocol) and the refit ($\woptp{P/F}{}'$ on 
projected activations).

Under $\mathrm{mp/raw}$, refit AUROC drops substantially:
mean $0.598$, range $0.516$--$0.680$ across the four models.
Two of four models (Qwen2.5-Instruct, Qwen3.5-0.8B) drop to
near-chance ($0.516$, $0.540$).
Under $\mathrm{lt/chat}$, refit AUROC retains more signal: 
mean $0.780$, range $0.719$--$0.813$, with all four models 
above $0.71$.
The diagnostic norm ratio
$\|\hat{\mu}^{\prime P/F}_\harm - \hat{\mu}^{\prime P/F}_\norm\| 
/ \|\hat{\mu}^{P/F}_\harm - \hat{\mu}^{P/F}_\norm\|$
on projected activations is below $10^{-3}$ across all
(model, protocol) pairs, confirming that the projection removes
the mean-difference component as designed.
The angle between baseline $\woptp{P/F}$ and refit 
$\woptp{P/F}{}'$ falls in the $80^\circ$--$89^\circ$ range 
across all conditions: refit $\woptp{P/F}{}'$ lies in the 
subspace orthogonal to $\wmdp{P/F}$ by construction, with the 
small residual angular deviation from $90^\circ$ accounted for 
by the angle between baseline $\woptp{mp/raw}$ and 
$\wmdp{mp/raw}$ (mean $9.1^\circ$ across the 12-model main 
set, Table~\ref{tab:angles_full}).

TPR@1\%FPR for refit conditions falls to zero with bootstrap CIs
of $[0, 0]$ in several cases (Table~\ref{tab:g4_refit}).
This reflects the reduced score separation under the strict
operating-point criterion: when AUROC drops to $0.5$--$0.7$, the
threshold at $1\%$ FPR sits above the maximum harmful score and
no harmful prompt clears it.

\begin{table}[ht]
\centering\small
\setlength{\tabcolsep}{4pt}
\caption{Within-protocol projection-and-refit on the four 
instruction-tuned models. The procedure runs independently 
under each of two protocols. \textbf{Protocols tested.} 
\emph{$\mathrm{mp/raw}$}: max-pool over content-token positions 
of raw prompts (the main-analysis protocol; $\wmdp{mp/raw}$ at 
$\Lp{mp/raw}$). \emph{$\mathrm{lt/chat}$}: last-token of 
chat-templated input ($\wmdp{lt/chat}$ at $\Lp{lt/chat}$). 
\textbf{Conditions per protocol.} \emph{baseline $\wmdp{P/F}$, 
baseline $\woptp{P/F}$}: AUROC on unprojected activations. 
\emph{$\woptp{P/F}{}'$ refit}: $\woptp{P/F}{}'$ trained from 
scratch on activations with $\wmdp{P/F}$ (the protocol's own 
mean-difference direction) projected out, $\mathbf{x}' = 
\mathbf{x} - (\mathbf{x} \cdot \wmdp{P/F})\,\wmdp{P/F}$. AUROC 
with stratified bootstrap 95\% CIs (1{,}000 resamples). 
In the Condition column, $\wmd$ and $\wopt$ refer to the protocol-specific direction in each row's protocol.}
\label{tab:g4_refit}
\begin{tabular}{llll}
\toprule
Model & Protocol & Condition & AUROC [95\% CI] \\
\midrule
  Qwen2.5-0.5B-Instruct & $\mathrm{mp/raw}$ & baseline $\wmd$ & 0.975 [0.967, 0.981] \\
  Qwen2.5-0.5B-Instruct & $\mathrm{mp/raw}$ & baseline $\wopt$ & 0.992 [0.989, 0.996] \\
  Qwen2.5-0.5B-Instruct & $\mathrm{mp/raw}$ & $\wopt'$ (refit, $\wmd$ projected out) & 0.516 [0.501, 0.537] \\
\midrule
  Qwen2.5-0.5B-Instruct & $\mathrm{lt/chat}$ & baseline $\wmd$ & 0.990 [0.986, 0.993] \\
  Qwen2.5-0.5B-Instruct & $\mathrm{lt/chat}$ & baseline $\wopt$ & 0.991 [0.987, 0.994] \\
  Qwen2.5-0.5B-Instruct & $\mathrm{lt/chat}$ & $\wopt'$ (refit, $\wmd$ projected out) & 0.719 [0.695, 0.746] \\
\midrule
  Qwen3.5-0.8B & $\mathrm{mp/raw}$ & baseline $\wmd$ & 0.964 [0.956, 0.972] \\
  Qwen3.5-0.8B & $\mathrm{mp/raw}$ & baseline $\wopt$ & 0.985 [0.979, 0.989] \\
  Qwen3.5-0.8B & $\mathrm{mp/raw}$ & $\wopt'$ (refit, $\wmd$ projected out) & 0.540 [0.518, 0.562] \\
\midrule
  Qwen3.5-0.8B & $\mathrm{lt/chat}$ & baseline $\wmd$ & 0.986 [0.982, 0.990] \\
  Qwen3.5-0.8B & $\mathrm{lt/chat}$ & baseline $\wopt$ & 0.988 [0.984, 0.992] \\
  Qwen3.5-0.8B & $\mathrm{lt/chat}$ & $\wopt'$ (refit, $\wmd$ projected out) & 0.802 [0.777, 0.824] \\
\midrule
  Llama-3.2-1B-Instruct & $\mathrm{mp/raw}$ & baseline $\wmd$ & 0.988 [0.982, 0.992] \\
  Llama-3.2-1B-Instruct & $\mathrm{mp/raw}$ & baseline $\wopt$ & 0.989 [0.984, 0.993] \\
  Llama-3.2-1B-Instruct & $\mathrm{mp/raw}$ & $\wopt'$ (refit, $\wmd$ projected out) & 0.657 [0.629, 0.682] \\
\midrule
  Llama-3.2-1B-Instruct & $\mathrm{lt/chat}$ & baseline $\wmd$ & 0.998 [0.997, 0.999] \\
  Llama-3.2-1B-Instruct & $\mathrm{lt/chat}$ & baseline $\wopt$ & 0.999 [0.998, 0.999] \\
  Llama-3.2-1B-Instruct & $\mathrm{lt/chat}$ & $\wopt'$ (refit, $\wmd$ projected out) & 0.813 [0.789, 0.837] \\
\midrule
  gemma-3-1b-it & $\mathrm{mp/raw}$ & baseline $\wmd$ & 0.957 [0.947, 0.966] \\
  gemma-3-1b-it & $\mathrm{mp/raw}$ & baseline $\wopt$ & 0.983 [0.977, 0.989] \\
  gemma-3-1b-it & $\mathrm{mp/raw}$ & $\wopt'$ (refit, $\wmd$ projected out) & 0.680 [0.650, 0.706] \\
\midrule
  gemma-3-1b-it & $\mathrm{lt/chat}$ & baseline $\wmd$ & 0.989 [0.986, 0.993] \\
  gemma-3-1b-it & $\mathrm{lt/chat}$ & baseline $\wopt$ & 0.994 [0.991, 0.996] \\
  gemma-3-1b-it & $\mathrm{lt/chat}$ & $\wopt'$ (refit, $\wmd$ projected out) & 0.788 [0.761, 0.812] \\
\bottomrule
\end{tabular}
\end{table}

\section{Cross-protocol projection-and-refit}
\label{app:cross_protocol_projection}

The previous appendix tests how concentrated the harm signal is
within one extraction protocol; here we test whether different
protocols recover the same concentrated direction or different
directions.
\textbf{Procedure.} At $\Lp{lt/chat}$ for each instruction-tuned 
model, we project $\wmdp{lt/chat}$ out of activations under 
three protocols: $\mathrm{lt/chat}$ (Arditi's own), 
$\mathrm{mp/chat}$, and $\mathrm{mp/raw}$ (the main-analysis 
protocol). We refit $\woptp{P/F}{}'$ on projected fit 
activations under each protocol $P/F$ and evaluate on projected 
eval activations.

Table~\ref{tab:g5_cross_protocol} reports baseline and refit
AUROC.
The pattern is consistent across the four models.
Under $\mathrm{lt/chat}$, projecting $\wmdp{lt/chat}$ drops 
AUROC substantially: mean change $-0.280$, range $-0.398$ to 
$-0.186$, with Llama-3.2-1B-Instruct showing the largest drop 
(0.999 to 0.601).
Under $\mathrm{mp/chat}$, the change is within bootstrap noise: 
mean $+0.006$, range $+0.001$ to $+0.020$.
Under $\mathrm{mp/raw}$, the change is also within bootstrap 
noise: mean $-0.002$, range $-0.010$ to $+0.001$.

The diagnostic norm ratio
$\|\hat{\mu}^{\prime P/F}_\harm - \hat{\mu}^{\prime P/F}_\norm\| 
/ \|\hat{\mu}^{P/F}_\harm - \hat{\mu}^{P/F}_\norm\|$
quantifies how much of the mean-difference vector under each
protocol $\wmdp{lt/chat}$ accounts for at $\Lp{lt/chat}$.
Under $\mathrm{lt/chat}$, the ratio is below $10^{-2}$ across 
all four models: $\wmdp{lt/chat}$ accounts for nearly all of 
the mean-difference vector at this protocol/layer, and 
projecting it out removes nearly all of the harm signal at this 
protocol.
Under $\mathrm{mp/chat}$, the ratio is $0.89$--$0.98$ across
models, indicating that $\wmdp{lt/chat}$ accounts for at most 
$11\%$ of the mean-difference vector under max-pool extraction 
of chat-templated input.
Under $\mathrm{mp/raw}$, the ratio is $0.99$ for all four 
models: $\wmdp{lt/chat}$ has essentially no presence in the 
mean-difference direction recovered by $\mathrm{mp/raw}$ 
extraction at the same layer.

The harm-signal concentrations recovered by max-pool and
last-token extraction at the same residual-stream layer are
near-orthogonal: removing one leaves the other essentially
intact.

\begin{table}[ht]
\centering\small
\setlength{\tabcolsep}{4pt}
\caption{Cross-protocol projection-and-refit at $\Lp{lt/chat}$. 
\textbf{Procedure.} For each instruction-tuned model and 
extraction protocol, project $\wmdp{lt/chat}$ out of 
activations, $\mathbf{x}' = \mathbf{x} - (\mathbf{x} \cdot 
\wmdp{lt/chat})\,\wmdp{lt/chat}$, and refit $\woptp{P/F}{}'$ on 
the projected fit set. \textbf{Protocols tested.} 
\emph{$\mathrm{lt/chat}$}: last-token of chat-templated input 
(Arditi's own protocol; baseline matches the diagonal of 
Table~\ref{tab:arditi_geometry}). \emph{$\mathrm{mp/chat}$}: 
max-pool over content-token positions of chat-templated input. 
\emph{$\mathrm{mp/raw}$}: max-pool over content-token positions 
of raw prompts (the main-analysis protocol). \textbf{Columns.} 
Baseline $\woptp{P/F}$: AUROC on unprojected activations. 
Refit $\woptp{P/F}{}'$: AUROC after the projection-and-refit. 
AUROC with stratified bootstrap 95\% CIs (1{,}000 resamples).}
\label{tab:g5_cross_protocol}
\begin{tabular}{llll}
\toprule
Model & Protocol & Baseline $\wopt$ & Refit $\wopt'$ after projecting $\wmdp{lt/chat}$ \\
\midrule
  Qwen2.5-0.5B-Instruct & $\mathrm{lt/chat}$ & 0.991 [0.987, 0.994] & 0.719 [0.695, 0.746] \\
  Qwen2.5-0.5B-Instruct & $\mathrm{mp/chat}$ & 0.929 [0.915, 0.941] & 0.949 [0.939, 0.958] \\
  Qwen2.5-0.5B-Instruct & $\mathrm{mp/raw}$ & 0.968 [0.960, 0.976] & 0.969 [0.961, 0.977] \\
\midrule
  Qwen3.5-0.8B & $\mathrm{lt/chat}$ & 0.988 [0.984, 0.992] & 0.802 [0.777, 0.824] \\
  Qwen3.5-0.8B & $\mathrm{mp/chat}$ & 0.990 [0.987, 0.993] & 0.991 [0.987, 0.993] \\
  Qwen3.5-0.8B & $\mathrm{mp/raw}$ & 0.982 [0.976, 0.988] & 0.982 [0.975, 0.987] \\
\midrule
  Llama-3.2-1B-Instruct & $\mathrm{lt/chat}$ & 0.999 [0.998, 0.999] & 0.601 [0.572, 0.630] \\
  Llama-3.2-1B-Instruct & $\mathrm{mp/chat}$ & 0.996 [0.994, 0.998] & 0.997 [0.996, 0.998] \\
  Llama-3.2-1B-Instruct & $\mathrm{mp/raw}$ & 0.981 [0.975, 0.988] & 0.982 [0.975, 0.988] \\
\midrule
  gemma-3-1b-it & $\mathrm{lt/chat}$ & 0.994 [0.991, 0.996] & 0.729 [0.701, 0.759] \\
  gemma-3-1b-it & $\mathrm{mp/chat}$ & 0.996 [0.994, 0.998] & 0.997 [0.995, 0.998] \\
  gemma-3-1b-it & $\mathrm{mp/raw}$ & 0.991 [0.988, 0.994] & 0.981 [0.976, 0.986] \\
\bottomrule
\end{tabular}
\end{table}

\section{Disaggregated out-of-distribution results}
\label{app:ood}

Tables~\ref{tab:ood_a} and~\ref{tab:ood_b} report per-model 
effective AUROC for each combination of harmful source 
(AdvBench, HarmBench, JailbreakBench) and benign source 
(Alpaca, XSTest), for $\wmdp{mp/raw}$ and $\woptp{mp/raw}$ at 
$\Lp{mp/raw}$. These disaggregate the model-averaged summary 
in Table~\ref{tab:ood} (main text), allowing inspection of 
per-model worst cases. AdvBench contributes to fitting; 
HarmBench, JailbreakBench, and XSTest are fully held out.

\begin{table}[ht]
\centering\footnotesize
\setlength{\tabcolsep}{3pt}
\caption{Disaggregated OOD AUROC, Part~1: Gemma-3 and Llama-3.2. Each cell: effective AUROC for the given harmful source vs benign source, at the validation-selected layer. AdvB: AdvBench, HB: HarmBench, JBB: JailbreakBench, Alp: Alpaca, XS: XSTest.}
\label{tab:ood_a}
\begin{tabular}{llrrrrrr}
\toprule
Model & Strategy & AdvB/Alp & AdvB/XS & HB/Alp & HB/XS & JBB/Alp & JBB/XS \\
\midrule
  \multirow{5}{*}{\shortstack[l]{Gemma-3\\(base)}} & Mean diff (LDA) & 0.985 & 0.993 & 0.942 & 0.954 & 0.940 & 0.948 \\
   & Soft-AUC & 0.987 & 0.994 & 0.943 & 0.954 & 0.941 & 0.948 \\
   & PC1 (normative) & 0.660 & 0.630 & 0.797 & 0.801 & 0.762 & 0.758 \\
   & $\theta$ normative & 0.501 & 0.758 & 0.682 & 0.530 & 0.621 & 0.596 \\
   & $\theta$ two-class & 0.977 & 0.987 & 0.939 & 0.948 & 0.935 & 0.940 \\
\midrule
  \multirow{5}{*}{\shortstack[l]{Gemma-3\\(instruct)}} & Mean diff (LDA) & 0.985 & 0.998 & 0.954 & 0.987 & 0.960 & 0.989 \\
   & Soft-AUC & 0.987 & 0.999 & 0.956 & 0.988 & 0.961 & 0.989 \\
   & PC1 (normative) & 0.634 & 0.724 & 0.773 & 0.849 & 0.750 & 0.828 \\
   & $\theta$ normative & 0.543 & 0.710 & 0.634 & 0.500 & 0.628 & 0.517 \\
   & $\theta$ two-class & 0.976 & 0.998 & 0.945 & 0.983 & 0.953 & 0.990 \\
\midrule
  \multirow{5}{*}{\shortstack[l]{Gemma-3\\(abliterated)}} & Mean diff (LDA) & 0.985 & 0.998 & 0.948 & 0.981 & 0.957 & 0.987 \\
   & Soft-AUC & 0.988 & 0.998 & 0.952 & 0.982 & 0.962 & 0.987 \\
   & PC1 (normative) & 0.587 & 0.690 & 0.742 & 0.830 & 0.717 & 0.811 \\
   & $\theta$ normative & 0.590 & 0.762 & 0.575 & 0.564 & 0.568 & 0.584 \\
   & $\theta$ two-class & 0.977 & 0.996 & 0.940 & 0.973 & 0.952 & 0.986 \\
\midrule
  \multirow{5}{*}{\shortstack[l]{Llama-3.2\\(base)}} & Mean diff (LDA) & 0.984 & 0.995 & 0.951 & 0.972 & 0.951 & 0.969 \\
   & Soft-AUC & 0.991 & 0.997 & 0.959 & 0.973 & 0.960 & 0.970 \\
   & PC1 (normative) & 0.630 & 0.806 & 0.780 & 0.909 & 0.747 & 0.892 \\
   & $\theta$ normative & 0.527 & 0.781 & 0.657 & 0.579 & 0.600 & 0.626 \\
   & $\theta$ two-class & 0.987 & 0.995 & 0.959 & 0.973 & 0.956 & 0.965 \\
\midrule
  \multirow{5}{*}{\shortstack[l]{Llama-3.2\\(instruct)}} & Mean diff (LDA) & 0.997 & 0.998 & 0.976 & 0.980 & 0.971 & 0.973 \\
   & Soft-AUC & 0.997 & 0.998 & 0.976 & 0.980 & 0.972 & 0.973 \\
   & PC1 (normative) & 0.583 & 0.631 & 0.768 & 0.806 & 0.737 & 0.781 \\
   & $\theta$ normative & 0.593 & 0.776 & 0.640 & 0.552 & 0.595 & 0.587 \\
   & $\theta$ two-class & 0.995 & 0.998 & 0.978 & 0.987 & 0.977 & 0.981 \\
\midrule
  \multirow{5}{*}{\shortstack[l]{Llama-3.2\\(abliterated)}} & Mean diff (LDA) & 0.996 & 0.999 & 0.979 & 0.983 & 0.975 & 0.978 \\
   & Soft-AUC & 0.997 & 0.999 & 0.982 & 0.985 & 0.978 & 0.980 \\
   & PC1 (normative) & 0.654 & 0.790 & 0.801 & 0.902 & 0.771 & 0.881 \\
   & $\theta$ normative & 0.508 & 0.748 & 0.660 & 0.581 & 0.630 & 0.605 \\
   & $\theta$ two-class & 0.995 & 0.998 & 0.979 & 0.984 & 0.975 & 0.980 \\
\bottomrule
\end{tabular}
\end{table}

\begin{table}[ht]
\centering\footnotesize
\setlength{\tabcolsep}{3pt}
\caption{Disaggregated OOD AUROC, Part~2: Qwen2.5 and Qwen3.5. Each cell: effective AUROC for the given harmful source vs benign source, at the validation-selected layer. AdvB: AdvBench, HB: HarmBench, JBB: JailbreakBench, Alp: Alpaca, XS: XSTest.}
\label{tab:ood_b}
\begin{tabular}{llrrrrrr}
\toprule
Model & Strategy & AdvB/Alp & AdvB/XS & HB/Alp & HB/XS & JBB/Alp & JBB/XS \\
\midrule
  \multirow{5}{*}{\shortstack[l]{Qwen2.5\\(base)}} & Mean diff (LDA) & 0.984 & 0.998 & 0.941 & 0.976 & 0.949 & 0.977 \\
   & Soft-AUC & 0.995 & 1.000 & 0.968 & 0.988 & 0.969 & 0.985 \\
   & PC1 (normative) & 0.582 & 0.588 & 0.750 & 0.660 & 0.718 & 0.613 \\
   & $\theta$ normative & 0.546 & 0.835 & 0.540 & 0.730 & 0.519 & 0.760 \\
   & $\theta$ two-class & 0.976 & 0.995 & 0.934 & 0.965 & 0.941 & 0.967 \\
\midrule
  \multirow{5}{*}{\shortstack[l]{Qwen2.5\\(instruct)}} & Mean diff (LDA) & 0.983 & 0.998 & 0.945 & 0.983 & 0.951 & 0.983 \\
   & Soft-AUC & 0.996 & 1.000 & 0.970 & 0.990 & 0.970 & 0.984 \\
   & PC1 (normative) & 0.575 & 0.672 & 0.745 & 0.595 & 0.716 & 0.547 \\
   & $\theta$ normative & 0.580 & 0.922 & 0.500 & 0.842 & 0.511 & 0.871 \\
   & $\theta$ two-class & 0.977 & 0.994 & 0.932 & 0.962 & 0.939 & 0.965 \\
\midrule
  \multirow{5}{*}{\shortstack[l]{Qwen2.5\\(abliterated)}} & Mean diff (LDA) & 0.970 & 0.996 & 0.922 & 0.971 & 0.926 & 0.968 \\
   & Soft-AUC & 0.992 & 0.999 & 0.957 & 0.982 & 0.953 & 0.976 \\
   & PC1 (normative) & 0.577 & 0.608 & 0.750 & 0.646 & 0.719 & 0.598 \\
   & $\theta$ normative & 0.617 & 0.917 & 0.511 & 0.829 & 0.557 & 0.882 \\
   & $\theta$ two-class & 0.958 & 0.986 & 0.904 & 0.941 & 0.907 & 0.938 \\
\midrule
  \multirow{5}{*}{\shortstack[l]{Qwen3.5\\(base)}} & Mean diff (LDA) & 0.993 & 0.999 & 0.969 & 0.989 & 0.975 & 0.990 \\
   & Soft-AUC & 0.993 & 0.999 & 0.970 & 0.989 & 0.976 & 0.990 \\
   & PC1 (normative) & 0.646 & 0.825 & 0.791 & 0.916 & 0.772 & 0.909 \\
   & $\theta$ normative & 0.582 & 0.814 & 0.554 & 0.679 & 0.537 & 0.706 \\
   & $\theta$ two-class & 0.991 & 0.999 & 0.971 & 0.994 & 0.980 & 0.994 \\
\midrule
  \multirow{5}{*}{\shortstack[l]{Qwen3.5\\(instruct)}} & Mean diff (LDA) & 0.984 & 0.995 & 0.926 & 0.948 & 0.936 & 0.944 \\
   & Soft-AUC & 0.994 & 0.998 & 0.954 & 0.968 & 0.956 & 0.957 \\
   & PC1 (normative) & 0.581 & 0.702 & 0.739 & 0.551 & 0.705 & 0.507 \\
   & $\theta$ normative & 0.625 & 0.970 & 0.690 & 0.957 & 0.691 & 0.965 \\
   & $\theta$ two-class & 0.978 & 0.951 & 0.880 & 0.743 & 0.906 & 0.808 \\
\midrule
  \multirow{5}{*}{\shortstack[l]{Qwen3.5\\(abliterated)}} & Mean diff (LDA) & 0.980 & 0.994 & 0.913 & 0.943 & 0.925 & 0.939 \\
   & Soft-AUC & 0.992 & 0.997 & 0.943 & 0.964 & 0.946 & 0.954 \\
   & PC1 (normative) & 0.568 & 0.678 & 0.732 & 0.579 & 0.696 & 0.530 \\
   & $\theta$ normative & 0.643 & 0.971 & 0.709 & 0.960 & 0.709 & 0.966 \\
   & $\theta$ two-class & 0.970 & 0.945 & 0.856 & 0.734 & 0.884 & 0.793 \\
\bottomrule
\end{tabular}
\end{table}

\paragraph{Worst-case OOD performance.}
For $\woptp{mp/raw}$, the in-distribution average AUROC 
(AdvBench vs.\ Alpaca) is 0.992; the worst model-averaged 
out-of-distribution cell is 0.961 (HarmBench vs.\ Alpaca); the 
worst single-model out-of-distribution cell across the 12 
models is 0.941 (Gemma-3 base, JailbreakBench vs.\ Alpaca). 
The average of all pairwise model-level out-of-distribution 
tests is 0.960.

\section{Full cross-variant transfer matrices}
\label{app:cross_variant}

Table~\ref{tab:transfer_full} reports the full $3 \times 3$
cross-variant transfer matrix for all four families. Each cell 
shows the raw AUROC (without sign correction) obtained by 
scoring one variant's evaluation data with another variant's 
$\wmdp{mp/raw}$ direction, fitted at the base model's 
validation-selected layer. Diagonal entries use each variant's 
own direction. All raw AUROC values exceed 0.5, confirming that 
no sign reversals occur in any cross-variant transfer.

\begin{table}[htb]
\centering\small
\caption{Full cross-variant transfer matrix: AUROC and TPR@1\%FPR.
  Each cell shows AUROC\,/\,TPR when scoring the target variant's
  evaluation data with the source variant's $\wmdp{mp/raw}$ direction,
  fitted at the base model's validation-selected layer.
  Diagonal entries (bold) use each variant's own direction.
  Italicised TPR values ($<0.25$) indicate near-zero operational
  detectability despite adequate AUROC\@.
  The lower block extends the analysis to Qwen3.5 at 2B, 4B, and 9B
  parameters; see Table~\ref{tab:transfer_scale} for the summary.
  $^\dagger$For Qwen3.5 (0.8B), layer\,10 (base-selected) outperforms the
  instruct and abliterated variants' own optimal layer\,(22).}
\label{tab:transfer_full}
\begin{tabular}{llcccccc}
\toprule
 & & \multicolumn{2}{c}{Base} & \multicolumn{2}{c}{Instruct} & \multicolumn{2}{c}{Ablit.} \\
\cmidrule(lr){3-4}\cmidrule(lr){5-6}\cmidrule(lr){7-8}
 & Source & AUC & TPR & AUC & TPR & AUC & TPR \\
\midrule
  \multirow{3}{*}{Llama-3.2} & base & \textbf{0.974} & \textbf{0.675} & 0.978 & 0.754 & 0.981 & 0.733 \\
   & instruct & 0.979 & 0.706 & \textbf{0.988} & \textbf{0.827} & 0.988 & 0.749 \\
   & abliterated & 0.978 & 0.600 & 0.985 & 0.679 & \textbf{0.988} & \textbf{0.701} \\
\midrule
  \multirow{3}{*}{Qwen2.5} & base & \textbf{0.973} & \textbf{0.648} & 0.973 & 0.597 & 0.958 & 0.479 \\
   & instruct & 0.974 & 0.685 & \textbf{0.975} & \textbf{0.643} & 0.963 & 0.516 \\
   & abliterated & 0.964 & 0.561 & 0.964 & 0.493 & \textbf{0.961} & \textbf{0.510} \\
\midrule
  \multirow{3}{*}{Qwen3.5$^\dagger$} & base & \textbf{0.987} & \textbf{0.818} & 0.988 & 0.815 & 0.986 & 0.767 \\
   & instruct & 0.982 & 0.757 & \textbf{0.987} & \textbf{0.821} & 0.985 & 0.797 \\
   & abliterated & 0.979 & 0.727 & 0.984 & 0.797 & \textbf{0.984} & \textbf{0.785} \\
\midrule
  \multirow{3}{*}{Gemma-3} & base & \textbf{0.969} & \textbf{0.706} & 0.922 & \textit{0.175} & 0.900 & \textit{0.100} \\
   & instruct & 0.888 & \textit{0.179} & \textbf{0.979} & \textbf{0.751} & 0.974 & 0.691 \\
   & abliterated & 0.899 & \textit{0.221} & 0.980 & 0.740 & \textbf{0.977} & \textbf{0.712} \\
\midrule
\multicolumn{8}{l}{Qwen3.5 scaling extension (same pipeline, larger sizes)} \\
\midrule
  \multirow{3}{*}{Qwen3.5-2B} & base & \textbf{0.986} & \textbf{0.794} & 0.990 & 0.873 & 0.988 & 0.842 \\
   & instruct & 0.990 & 0.830 & \textbf{0.993} & \textbf{0.870} & 0.991 & 0.843 \\
   & abliterated & 0.990 & 0.821 & 0.992 & 0.860 & \textbf{0.991} & \textbf{0.855} \\
\midrule
  \multirow{3}{*}{Qwen3.5-4B} & base & \textbf{0.979} & \textbf{0.716} & 0.986 & 0.813 & 0.984 & 0.767 \\
   & instruct & 0.990 & 0.788 & \textbf{0.994} & \textbf{0.907} & 0.993 & 0.878 \\
   & abliterated & 0.982 & 0.703 & 0.989 & 0.763 & \textbf{0.991} & \textbf{0.822} \\
\midrule
  \multirow{3}{*}{Qwen3.5-9B} & base & \textbf{0.992} & \textbf{0.858} & 0.994 & 0.875 & 0.991 & 0.836 \\
   & instruct & 0.994 & 0.867 & \textbf{0.995} & \textbf{0.893} & 0.993 & 0.855 \\
   & abliterated & 0.992 & 0.828 & 0.993 & 0.852 & \textbf{0.991} & \textbf{0.846} \\
\bottomrule
\end{tabular}
\end{table}

\paragraph{Asymmetry of Gemma-3 transfer.}
The Gemma-3 transfer is asymmetric: the base direction applied 
to instruct data produces an AUROC of $0.922$, while the 
instruct direction applied to base data achieves $0.889$ 
(bootstrap 95\% CIs within $\pm 0.007$). Fitted directions are 
stochastic estimates from 100 prompts per class, and the chosen 
direction need not coincide with the eval-set optimum, so the 
asymmetry admits multiple explanations including statistical 
noise, distributional change in the class manifolds under 
instruction tuning, and direction-level changes induced by 
alignment.

\section{Scale stability}
\label{app:scale}

We replicate the cross-variant analysis on Qwen3.5 at 2B, 4B, 
and 9B parameters (Table~\ref{tab:transfer_scale}). Across 
parameter scales from 0.8B to 9B in the Qwen3.5 family, AUROC 
remains in $[0.98, 0.99]$ on all variants under 
$\mathrm{mp/raw}$, cross-variant transfer remains within 0.018 
AUROC of own-direction performance, and pairwise direction 
angles remain in the $11^\circ$--$27^\circ$ range. The 9B 
variant exhibits the cleanest cross-variant transfer of any 
model in our experimental set: every off-diagonal AUROC is 
within 0.004 of the corresponding diagonal.

\begin{table}[htb]
\centering\small
\caption{Cross-variant direction transfer across Qwen3.5 model sizes.
  Focused on the
  Qwen3.5 family across an 11$\times$ parameter range (0.8B--9B).
  AUROC, TPR@1\%FPR and direction-angle conventions are as in
  Table~\ref{tab:cross_variant}.}
\label{tab:transfer_scale}
\begin{tabular}{lcccccccc}
\toprule
 & \multicolumn{3}{c}{Angles}
   & \multicolumn{2}{c}{Instruct}
   & \multicolumn{2}{c}{Abliterated} \\
\cmidrule(lr){2-4} \cmidrule(lr){5-6} \cmidrule(lr){7-8}
Size & B$\leftrightarrow$I & B$\leftrightarrow$A & I$\leftrightarrow$A
       & AUROC & TPR@1\%FPR
       & AUROC & TPR@1\%FPR \\
\midrule
0.8B & $17^\circ$ & $20^\circ$ & $11^\circ$ & 0.988 ($+$0.001) & 0.815 ($-$0.006) & 0.986 ($+$0.002) & 0.767 ($-$0.018) \\
2B & $18^\circ$ & $21^\circ$ & $12^\circ$ & 0.990 ($-$0.002) & 0.873 ($+$0.003) & 0.988 ($-$0.003) & 0.842 ($-$0.013) \\
4B & $17^\circ$ & $27^\circ$ & $24^\circ$ & 0.986 ($-$0.008) & 0.813 ($-$0.094) & 0.984 ($-$0.007) & 0.767 ($-$0.055) \\
9B & $16^\circ$ & $20^\circ$ & $13^\circ$ & 0.994 ($-$0.001) & 0.875 ($-$0.018) & 0.991 ($\pm$0.000) & 0.836 ($-$0.010) \\
\bottomrule
\end{tabular}
\end{table}

\section{Sample efficiency}
\label{app:sample_efficiency}

\begin{figure}[ht]
\centering
\includegraphics[width=\textwidth]{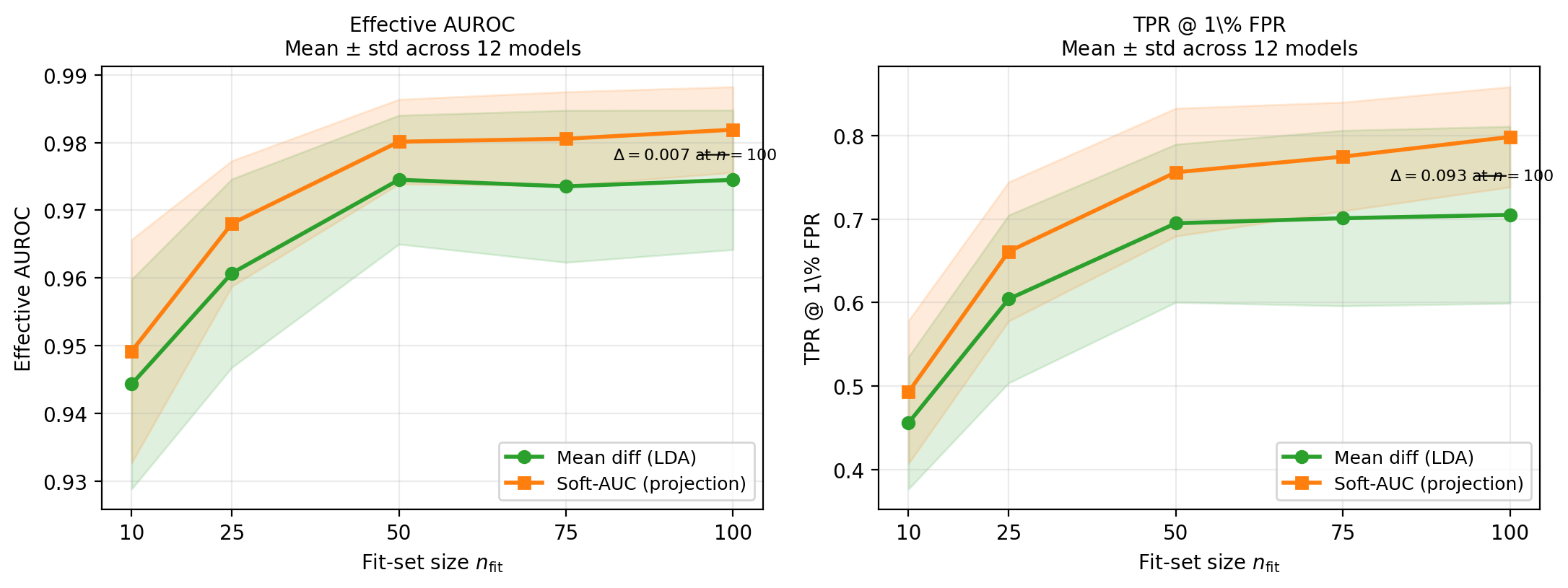}
\caption{Effective AUROC (left) and TPR@1\%FPR (right) as a
  function of fit-set size $n_\mathrm{fit}$, averaged across all
  12 models ($\pm$1~std shaded), under $\mathrm{mp/raw}$ 
  extraction. Mean over 5 random subsamples per $n$.}
\label{fig:sample_eff}
\end{figure}

Figure~\ref{fig:sample_eff} reports AUROC and TPR@1\%FPR as a
function of fit-set size $n \in \{10, 25, 50, 75, 100\}$ under 
$\mathrm{mp/raw}$ extraction, averaged over 5 random subsamples 
per $n$ and across all 12 models. $\wmdp{mp/raw}$ plateaus 
near $n = 50$ for both metrics. $\woptp{mp/raw}$ continues to 
improve through $n = 100$, with $\Delta\mathrm{AUROC} = 0.007$ 
and $\Delta\mathrm{TPR} = 0.093$ over $\wmdp{mp/raw}$ at 
$n = 100$.

\section{Safety classifier benchmark}
\label{app:classifier_comparison}

We benchmark four publicly available safety classifiers on the 
same evaluation set used in our main analysis: Llama 
Guard~3~\citep{dubey2024llama3herdmodels}, 
WildGuard~\citep{han2024wildguard}, 
ShieldGemma~\citep{zeng2024shieldgemma}, and Latent 
Guard~\citep{zhao2025llms}. Table~\ref{tab:classifiers} reports 
headline AUROC, TPR@1\%FPR, and end-to-end inference latency. 
Table~\ref{tab:classifier_full} reports per-source AUROC and 
TPR@1\%FPR.

\begin{table}[ht]
\centering\small
\caption{Comparison with dedicated safety classifiers, 
  evaluated on the same held-out splits used in our main 
  analysis. Standalone latency is the full pipeline cost 
  including the forward pass. Marginal latency is the cost 
  when the probe shares a forward pass with an existing 
  inference. Hardware: classifiers benchmarked on RTX~3090; 
  our method on RTX~3070 Mobile.}
\label{tab:classifiers}
\begin{tabular}{lccccc}
\toprule
Method & Type & Params & AUROC & TPR@1\%FPR & Latency (ms) \\
\midrule
$\woptp{mp/raw}$ (ours)            & probe        & host    & 0.982 & 0.797 & 4 / 26$^\dagger$ \\
Latent Guard~\citep{zhao2025llms}      & probe        & host    & 0.959 & 0.576 & --- / 16$^\dagger$ \\
\midrule
Llama Guard~3~\citep{dubey2024llama3herdmodels}     & classifier   & 8B      & 0.994 & 0.932 & 74 \\
WildGuard~\citep{han2024wildguard}        & classifier   & 7B      & 0.998 & 0.996 & 50 \\
ShieldGemma~\citep{zeng2024shieldgemma}      & classifier   & 9B      & 0.939 & 0.304 & 90 \\
\bottomrule
\end{tabular}
\smallskip

$^\dagger$Marginal / standalone.
\end{table}

\paragraph{Latency.}
Trained classifiers require an independent forward pass through 
a 7--9B model (50--90~ms on RTX~3090). Activation-based probes 
can run standalone (26~ms for ours, 16~ms for Latent Guard) or 
share a forward pass with an existing inference, in which case 
the marginal cost reduces to the activation hook and scoring 
step (4~ms).

\paragraph{Patterns in the disaggregated results.}
WildGuard achieves near-perfect performance across all
combinations of harm source and benign source 
(AUROC $\geq 0.998$, TPR $\geq 0.975$). ShieldGemma exhibits 
substantial AUROC--TPR pattern on hard-benign XSTest: TPR drops 
from 0.80--0.95 (vs.\ Alpaca) to 0.15--0.31 (vs.\ XSTest), while 
AUROC drops from 0.99 to 0.76--0.86.

\begin{table}[ht]
\centering
\caption{Per-source AUROC and TPR@1\%FPR for each safety
  classifier evaluated on our held-out splits.}
\label{tab:classifier_full}
\begin{tabular}{llrrrr}
\toprule
Method & Harm source & vs.\ Alpaca AUROC & TPR & vs.\ XSTest AUROC & TPR \\
\midrule
$\woptp{mp/raw}$ (ours, mean)
  & AdvBench       & 0.992 & 0.867 & 0.996 & 0.941 \\
  & HarmBench      & 0.961 & 0.596 & 0.973 & 0.769 \\
  & JailbreakBench & 0.963 & 0.650 & 0.970 & 0.807 \\
\midrule
Llama Guard~3
  & AdvBench       & 0.998 & 0.985 & 0.995 & 0.925 \\
  & HarmBench      & 0.992 & 0.944 & 0.983 & 0.869 \\
  & JailbreakBench & 0.994 & 0.962 & 0.985 & 0.876 \\
\midrule
WildGuard
  & AdvBench       & 0.998 & 0.997 & 0.999 & 0.992 \\
  & HarmBench      & 0.998 & 1.000 & 0.999 & 0.990 \\
  & JailbreakBench & 0.998 & 1.000 & 0.999 & 0.975 \\
\midrule
ShieldGemma
  & AdvBench       & 0.997 & 0.954 & 0.856 & 0.264 \\
  & HarmBench      & 0.990 & 0.845 & 0.816 & 0.305 \\
  & JailbreakBench & 0.988 & 0.800 & 0.762 & 0.145 \\
\midrule
Latent Guard
  & AdvBench       & 0.988 & 0.757 & 0.967 & 0.605 \\
  & HarmBench      & 0.950 & 0.491 & 0.890 & 0.310 \\
  & JailbreakBench & 0.954 & 0.540 & 0.901 & 0.400 \\
\bottomrule
\end{tabular}
\end{table}

\section{On the non-uniqueness of fitted linear directions}
\label{app:linearity}

Section~\ref{sec:discussion} notes that multiple non-collinear directions can support comparable AUROC, and that our procedure recovers one such direction.

For a binary classification problem with class means
$\mu_+, \mu_-$ and shared within-class covariance $\Sigma$, the
Fisher Linear Discriminant direction is
$\mathbf{w}_{\mathrm{FLD}} \propto \Sigma^{-1}(\mu_+ - \mu_-)$.
Under the equal-spherical-covariance assumption ($\Sigma =
\sigma^2 \mathbf{I}$), this reduces to $\wmd$
(Equation~\ref{eq:mean_diff}). The classification rule
$\mathbf{1}[\mathbf{w} \cdot \mathbf{x} > t]$ has the same
decision boundary for any $\mathbf{w}'$ collinear with
$\mathbf{w}$, so collinear directions are AUROC-equivalent. More 
generally, any $\mathbf{w}'$ in the affine span of vectors that 
achieve a given separating hyperplane on the empirical 
distribution will yield comparable AUROC.

The empirical observations support this directly. 
$\wmdp{mp/raw}$ and $\woptp{mp/raw}$ lie at 
$9.1^\circ \pm 8.8^\circ$ across models: close, but not 
collinear (Table~\ref{tab:angles_full}). $\woptp{mp/chat}$ and 
$\wmdp{lt/chat}$, both mean-difference-derived directions on 
chat-templated activations, lie at $72.6^\circ \pm 6.7^\circ$ 
across instruction-tuned models 
(Table~\ref{tab:arditi_geometry}), substantially further apart 
yet supporting near-equivalent AUROC on their respective 
protocols. Refitting in the subspace orthogonal to 
$\wmdp{lt/chat}$ at $\Lp{lt/chat}$ recovers AUROC within 
bootstrap noise of the unprojected baseline on three of four 
instruction-tuned models and within $0.020$ on the fourth 
(Qwen2.5; Appendix~\ref{app:projection_test}). Several 
non-collinear mean-difference directions, distinguished by 
extraction protocol and validation criterion, all serve as 
viable scoring axes for the same hyperplane. Detection-level 
evidence identifies the recoverability of such a direction, 
not the specific direction the model uses computationally.

\end{document}